\documentclass[11pt,a4paper]{article}
\usepackage{times,latexsym}
\usepackage{url}
\usepackage[T1]{fontenc}
\usepackage{CJKutf8}

\usepackage[acceptedWithA]{tacl2018v2}
\usepackage[]{tacl2018v2}

\usepackage{graphicx}
\usepackage{bm}
\usepackage{amssymb}
\usepackage{amsmath}
\usepackage[T1]{fontenc}
\usepackage{booktabs}
\usepackage{xcolor}
\usepackage{multirow}
\usepackage{booktabs}
\usepackage{verbatim}
\usepackage{mdwlist}
\usepackage[ruled,norelsize,linesnumbered]{algorithm2e}
\usepackage{amsthm}
\usepackage[capitalize]{cleveref}
\usepackage{paralist}
\usepackage{enumitem}
\usepackage{subcaption}

\theoremstyle{definition}

\newcommand{\e}[2]{\mathbb{E}_{#1}\left[ #2 \right] }

\newcommand{\kld}[2]{D_{\mathrm{KL}} \left[ \left. \left. #1 \right|\right| #2 \right] }

\makeatletter
\newcommand{\removelatexerror}{\let\@latex@error\@gobble}
\makeatother

\usepackage{paralist}
\setitemize{itemsep=1pt,topsep=1pt,parsep=1pt,partopsep=1pt}

\usepackage{xspace}
\newcommand{\modelname}{EDITOR\xspace}
\newcommand{\ar}{AR\xspace}
\newcommand{\nar}{NAR\xspace}
\newcommand{\levt}{LevT\xspace}

\title{\modelname: an Edit-Based Transformer with Repositioning \\ for Neural Machine Translation with Soft Lexical Constraints}

\author{Weijia Xu \\
	University of Maryland \\
	{\tt \href{mailto:weijia@cs.umd.edu}{weijia@cs.umd.edu}} \\\And
	Marine Carpuat \\
	University of Maryland \\
	{\tt \href{mailto:marine@cs.umd.edu}{marine@cs.umd.edu}} \\}

\date{}

\begin{document}

\maketitle
\begin{abstract}
We introduce an \textbf{Edi}t-Based \textbf{T}ransf\textbf{O}rmer with \textbf{R}epositioning (EDITOR), which makes sequence generation flexible by seamlessly allowing users to specify preferences in output lexical choice. Building on recent models for non-autoregressive sequence generation~\citep{GuWZ2019}, \modelname generates new sequences by iteratively editing hypotheses. It relies on a novel reposition operation designed to disentangle lexical choice from word positioning decisions, while enabling efficient oracles for imitation learning and parallel edits at decoding time. Empirically, \modelname uses soft lexical constraints more effectively than the Levenshtein Transformer~\citep{GuWZ2019} while speeding up decoding dramatically compared to constrained beam search~\citep{PostV2018}.
\modelname also achieves comparable or better translation quality with faster decoding speed than the Levenshtein Transformer on standard Romanian-English, English-German, and English-Japanese machine translation tasks.
\end{abstract}

\section{Introduction}
\looseness=-1
Neural machine translation (MT) architectures~\citep{BahdanauCB15,Vaswani2017} make it difficult for users to specify preferences that could be incorporated more easily in statistical MT models \citep{Koehn2007Moses} and have been shown to be useful for interactive machine translation~\citep{Foster2002,Barrachina2009} and domain adaptation~\citep{HokampL2017}. Lexical constraints or preferences have previously been incorporated by re-training NMT models with constraints as inputs~\citep{SongZYLWZ2019,DinuMFA2019} or with constrained beam search that drastically slows down decoding~\citep{HokampL2017,PostV2018}.

\begin{figure}[!t]
    \centering
    \includegraphics[width=0.48\textwidth]{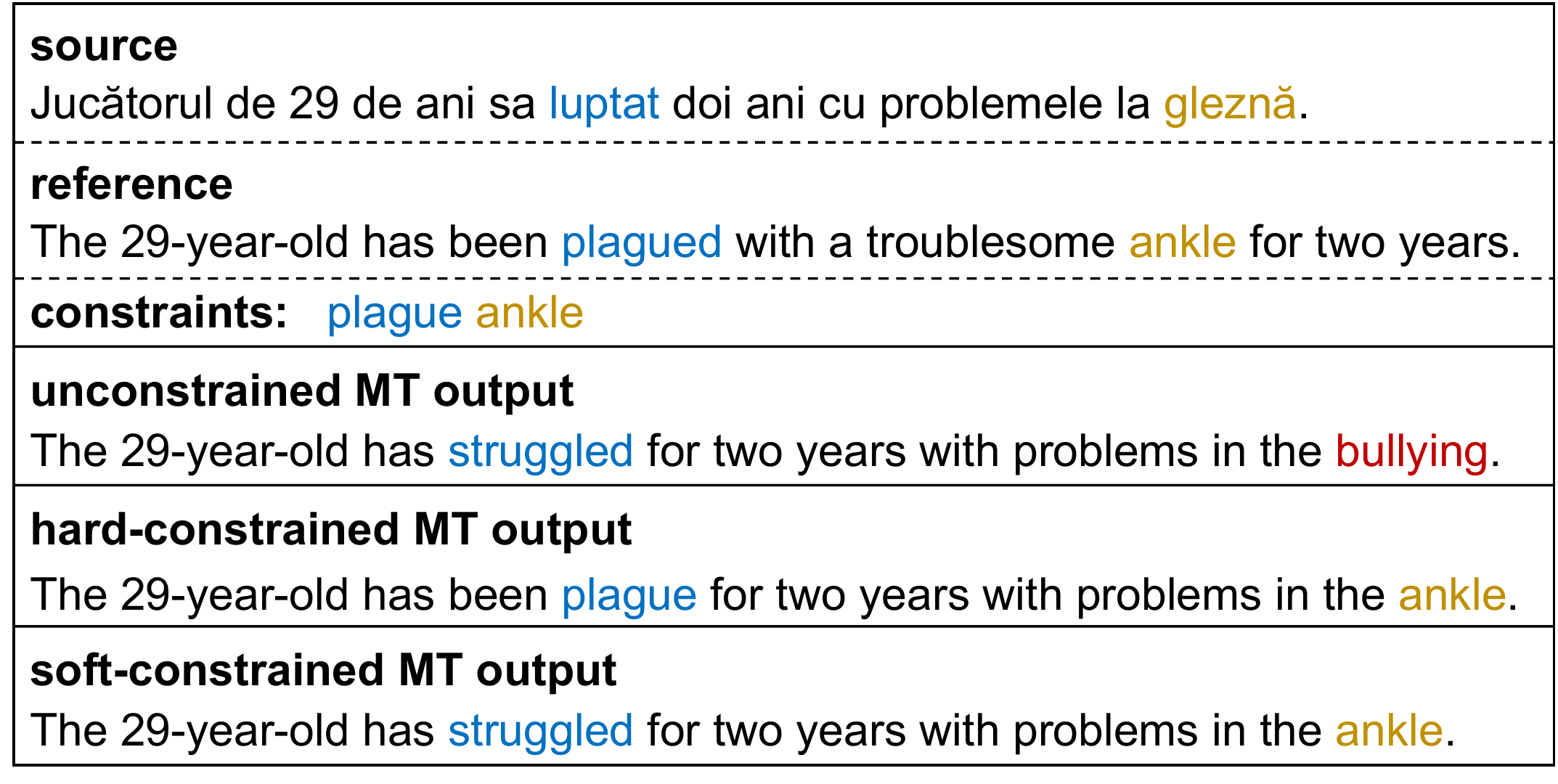}
    \caption{Romanian to English MT example. Unconstrained MT incorrectly translates ``gleznă'' to ``bullying''. Given constraint words ``plague'' and ``ankle'', soft-constrained MT correctly uses ``ankle'' and avoids disfluencies introduced by using ``plague'' as a hard constraint in its exact form.}
\label{fig:mt_example}
\end{figure}

\looseness=-1
In this work, we introduce a translation model that can seamlessly incorporate users' lexical choice preferences without increasing the time and computational cost at decoding time, while being trained on regular MT samples. We apply this model to MT tasks with soft lexical constraints. As illustrated in Figure~\ref{fig:mt_example}, when decoding with soft lexical constraints, user preferences for lexical choice in the output language are provided as an additional input sequence of target words in any order. The goal is to let users encode terminology, domain or stylistic preferences in target word usage, without strictly enforcing hard constraints that might hamper NMT's ability to generate fluent outputs.

Our model is an \textbf{Edi}t-Based \textbf{T}ransf\textbf{O}rmer with \textbf{R}epositioning (\textbf{EDITOR}), which builds on recent progress on non-autoregressive sequence generation~\citep{LeeMC2018,GhazvininejadLLZ2019}.\footnote{\url{https://github.com/Izecson/fairseq-editor}} Specifically, the Levenshtein Transformer~\citep{GuWZ2019} showed that iteratively refining output sequences via insertions and deletions yields a fast and flexible generation process for MT and automatic post-editing tasks. \modelname replaces the deletion operation with a novel reposition operation to disentangle lexical choice from reordering decisions. As a result, \modelname exploits lexical constraints more effectively and efficiently than the Levenshtein Transformer, as a single reposition operation can subsume a sequence of deletions and insertions. To train \modelname via imitation learning, the reposition operation is defined to preserve the ability to use the Levenshtein edit distance~\citep{Levenshtein1966} as an efficient oracle. We also introduce a dual-path roll-in policy which lets the reposition and deletion models learn to refine their respective outputs more effectively.

\looseness=-1
Experiments on Romanian-English, English-German, and English-Japanese MT show that \modelname achieves comparable or better translation quality with faster decoding speed than the Levenshtein Transformer \citep{GuWZ2019} on the standard MT tasks and exploit soft lexical constraints better: it achieves significantly better translation quality and matches more constraints with faster decoding speed than the Levenshtein Transformer. It also drastically speeds up decoding compared to lexically constrained decoding algorithms~\citep{PostV2018}. Furthermore, results highlight the benefits of soft constraints over hard ones \---\ \modelname with soft constraints achieves translation quality on par or better than both \modelname and Levenshtein Transformer with hard constraints~\citep{SusantoCT2020}.

\section{Background}

\paragraph{Non-Autoregressive MT} 
While autoregressive models that decode from left-to-right are the \textit{de facto} standard for many sequence generation tasks \citep{Cho2014,Chorowski2015,Vinyals2015}, non-autoregressive models offer a promising alternative to speed up decoding by generating a sequence of tokens in parallel~\citep{GuBXLS2018,Oord2018,MaZLNH2019}. However, their output quality suffers due to the large decoding space and strong independence assumptions between target tokens~\citep{MaZLNH2019,WangTHQZL2019}. These issues have been addressed via partially parallel decoding~\citep{WangZC2018,SternSU2018} or multi-pass decoding~\citep{LeeMC2018,GhazvininejadLLZ2019,GuWZ2019}. This work adopts multi-pass decoding, where the model generates the target sequences by iteratively editing the outputs from previous iterations. Edit operations such as substitution~\citep{GhazvininejadLLZ2019} and insertion-deletion~\citep{GuWZ2019} have reduced the quality gap between non-autoregressive and autoregressive models.
However, we argue that these operations limit the flexibility and efficiency of the resulting models for MT by entangling lexical choice and reordering decisions.

\paragraph{Reordering vs. Lexical Choice} \modelname's insertion and reposition operations connect closely with the long-standing view of MT as a combination of a translation or lexical choice model \---\ which selects appropriate translations for source units given their context \---\ and reordering model \---\ which encourages the generation of a target sequence order appropriate for the target language. This view is reflected in architectures ranging from the word-based IBM models~\citep{Brown1990SMT}, sentence-level models that generate a bag of target words that is reordered to construct a target sentence~\citep{BangaloreHK2007}, or the Operation Sequence Model~\citep{DurraniSFKS2015,StahlbergSB2018}, which views translation as a sequence of translation and reordering operations over bilingual minimal units. By contrast, autoregressive NMT models~\citep{BahdanauCB15,Vaswani2017} do not explicitly separate lexical choice and reordering, and previous non-autoregressive models break up reordering into sequences of other operations. This work introduces the reposition operation which makes it possible to move words around during the refinement process, as reordering models do. However, we will see that reposition differs from typical reordering to enable efficient oracles for training via imitation learning, and parallelization of edit operations at decoding time (Section~\ref{sec:approach}). 

\paragraph{MT with Soft Lexical Constraints} NMT models lack flexible mechanisms to incorporate users preferences in their outputs. 
Lexical constraints have been incorporated in prior work via  
\begin{inparaenum}[1)]
    \item constrained training where NMT models are trained on parallel samples augmented with constraint target phrases in both the source and target sequences~\citep{SongZYLWZ2019,DinuMFA2019}, or
    \item constrained decoding where beam search is modified to include constraint words or phrases in the output~\citep{HokampL2017,PostV2018}.
\end{inparaenum}
These mechanisms can incorporate domain-specific knowledge and lexicons which is particularly helpful in low-resource cases~\citep{Arthur2016,Tang2016}. Despite their success at domain adaptation for MT~\citep{HokampL2017} and caption generation~\citep{AndersonFJG2017}, they suffer from several issues: constrained training requires building dedicated models for constrained language generation, while constrained decoding adds significant computational overhead and treats all constraints as hard constraints which may hurt fluency.
In other tasks, various constraint types have been introduced by designing complex architectures tailored to specific content or style constraints~\citep{AbuI2011,MeiBW2016}, or via segment-level ``side-constraints''~\citep{SennrichHB2016politedness,FiclerG2017,AgrawalC2019}, which condition generation on users' stylistic preferences, but do not offer fine-grained control over their realization in the output sequence. 
We refer the reader to \citet{YvonS2020} for a comprehensive review of the strengths and weaknesses of current techniques to incorporate terminology constraints in NMT.

Our work is closely related to \citet{SusantoCT2020}'s idea of applying the Levenshtein Transformer to MT with hard terminology constraints. We will see that their technique can directly be used by \modelname as well (Section~\ref{sec:approach_inference}), but this does not offer empirical benefits over the default \modelname model~(Section~\ref{sec:lcmt_exp}).

\section{Approach}
\label{sec:approach}
\subsection{The \modelname Model}
\label{sec:basic_operations}

We cast both constrained and unconstrained language generation as an iterative sequence refinement problem modeled by a Markov Decision Process~$(\mathcal{Y}, \mathcal{A}, \mathcal{E}, \mathcal{R}, \boldsymbol{y}^0)$, where a state~$\boldsymbol{y}$ in the state space~$\mathcal{Y}$ corresponds to a sequence of tokens~$\boldsymbol{y} = (y_1, y_2, ..., y_L)$ from the vocabulary~$\mathcal{V}$ up to length~$L$, and~$\boldsymbol{y}^0 \in \mathcal{Y}$ is the initial sequence For standard sequence generation tasks, $\boldsymbol{y}^0$ is the empty sequence~$(\langle s \rangle, \langle /s \rangle)$. For lexically constrained generation tasks, $\boldsymbol{y}^0$ consists of the words to be used as constraints~$(\langle s \rangle, c_1, ..., c_m, \langle /s \rangle)$.

At the~$k$-th decoding iteration, the model takes as input $\boldsymbol{y}^{k - 1}$, the output from the previous iteration, chooses an action~$\boldsymbol{a}^k \in \mathcal{A}$ to refine the sequence into~$\boldsymbol{y}^k = \mathcal{E}(\boldsymbol{y}^{k - 1}, \boldsymbol{a}^k)$, and receives a reward~$r^k = \mathcal{R}(\boldsymbol{y}^k)$. The policy~$\pi$ maps the input sequence~$\boldsymbol{y}^{k - 1}$ to a probability distribution~$P(\mathcal{A})$ over the action space~$\mathcal{A}$. 
Our model is based on the Transformer encoder-decoder~\cite{Vaswani2017} and we extract the decoder representations~$(\boldsymbol{h}_1, ..., \boldsymbol{h}_n)$ to make the policy predictions.
Each refinement action is based on two basic operations: reposition and insertion. 

\begin{figure}[!t]
    \centering
    \includegraphics[width=.45\textwidth]{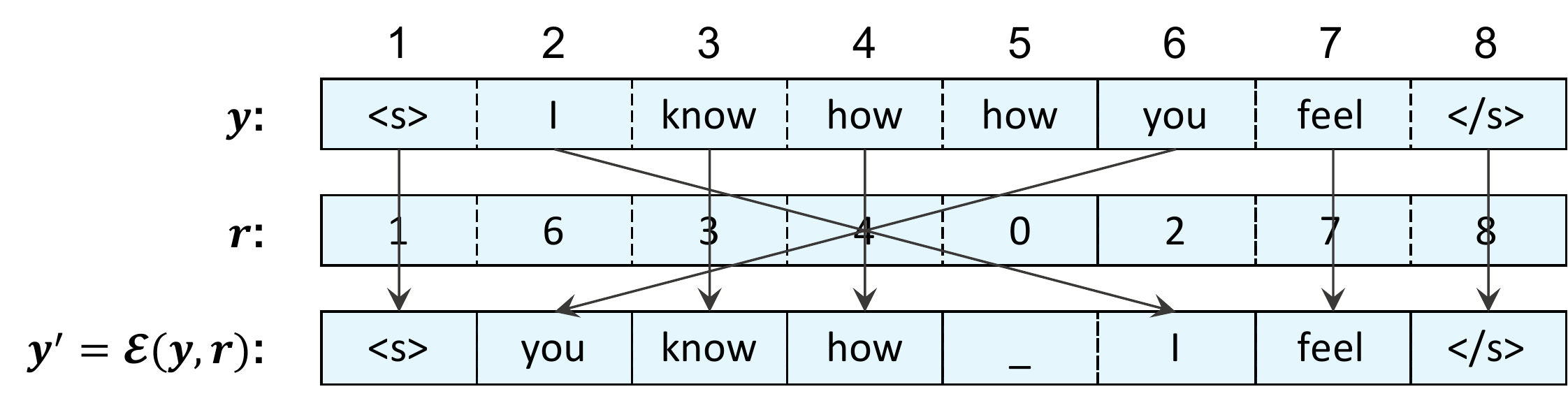}
\caption{Applying the reposition operation $\boldsymbol{r}$ to input $\boldsymbol{y}$: $r_i > 0$ is the 1-based index of token $y_i'$ in the input sequence; $y_i$ is deleted if $r_i =0$.}
\label{fig:reposition_example}
\end{figure}

\paragraph{Reposition}
\looseness=-1
For each position $i$ in the input sequence~$\boldsymbol{y}_{1...n}$, the reposition policy~$\pi_{rps}(r \,|\, i, \boldsymbol{y})$ predicts an index~$r \in [0, n]$: if~$r > 0$, we place the $r$-th input token~$y_r$ at the~$i$-th output position, otherwise we delete the token at that position~(Figure~\ref{fig:reposition_example}).  We constrain~$\pi_{rps}(1 \,|\, 1, \boldsymbol{y}) = \pi_{rps}(n \,|\, n, \boldsymbol{y}) = 1$ to maintain sequence boundaries. Note that reposition differs from typical reordering since
\begin{inparaenum}[1)]
    \item it makes it possible to delete tokens, and
    \item it places tokens at each position independently, which enables parallelization at decoding time. In principle, the same input token can thus be placed at multiple output positions. However, this happens rarely in practice as the policy predictor is trained to follow oracle demonstrations which cannot contain such repetitions by design.\footnote{Empirically, fewer than 1\% of tokens are repositioned to more than one output position.}
\end{inparaenum}

The reposition classifier gives a categorical distribution over the index of the input token to be placed at each output position:
\begin{equation}
\label{eq:rps_classifier}
    \pi_{rps}(r \,|\, i, \boldsymbol{y}) = \text{softmax}(\boldsymbol{h}_i \cdot [\boldsymbol{b}, \boldsymbol{e}_1, ..., \boldsymbol{e}_n])
\end{equation}
where~$\boldsymbol{e}_j$ is the embedding of the~$j$-th token in the input sequence, and~$\boldsymbol{b} \in \mathbb{R}^{d_{model}}$ is used to predict whether to delete the token. The dot product in the softmax function captures the similarity between the hidden state~$\boldsymbol{h}_i$ and each input embedding~$\boldsymbol{e}_j$ or the deletion vector~$\boldsymbol{b}$.

\paragraph{Insertion} 
Following \newcite{GuWZ2019}, the insertion operation consists of two phases:~(1) \textit{placeholder insertion:} given an input sequence~$\boldsymbol{y}_{1...n}$, the placeholder predictor~$\pi_{plh}(p \,|\, i, \boldsymbol{y})$ predicts the number of placeholders~$p \in [0, K_{max}]$ to be inserted between two neighboring tokens~$(y_i, y_{i + 1})$;\footnote{In our implementation, we set~$K_{max} = 255$.}~(2) \textit{token prediction:} given the output of the placeholder predictor, the token predictor~$\pi_{tok}(t \,|\, i, \boldsymbol{y})$ replaces each placeholder with an actual token.

The Placeholder Insertion Classifier gives a categorical distribution over the number of placeholders to be inserted between every two consecutive positions:
\begin{equation}
\label{eq:plh_classifier}
    \pi_{plh}(p \,|\, i, \boldsymbol{y}) = \text{softmax}([\boldsymbol{h}_i \,;\, \boldsymbol{h}_{i + 1}] \cdot \boldsymbol{W}^{plh})
\end{equation}
where~$\boldsymbol{W}^{plh} \in \mathbb{R}^{(2d_{model}) \times (K_{max} + 1)}$.

The Token Prediction Classifier predicts the identity of each token to fill in each placeholder:
\begin{equation}
\label{eq:tok_classifier}
    \pi_{tok}(t \,|\, i, \boldsymbol{y}) = \text{softmax}(\boldsymbol{h}_i \cdot \boldsymbol{W}^{tok})
\end{equation}
where~$\boldsymbol{W}^{tok} \in \mathbb{R}^{d_{model} \times |\mathcal{V}|}$.

\paragraph{Action} Given an input sequence~$\boldsymbol{y}_{1...n}$, an action consists of repositioning tokens, inserting and replacing placeholders. Formally, we define an action as \textit{a sequence} of reposition~($\boldsymbol{r}$), placeholder insertion~($\boldsymbol{p}$), and token prediction~($\boldsymbol{t}$) operations: $\boldsymbol{a} = (\boldsymbol{r}, \boldsymbol{p}, \boldsymbol{t})$. $\boldsymbol{r}$,~$\boldsymbol{p}$, and $\boldsymbol{t}$ are applied in this order to adjust non-empty initial sequences via reposition before inserting new tokens. Each of $\boldsymbol{r}$,~$\boldsymbol{p}$, and~$\boldsymbol{t}$ consists of a set of basic operations that can be applied \textit{in parallel}:
\[
\begin{split}
& \boldsymbol{r} = \{ r_1, ..., r_n \} \\
& \boldsymbol{p} = \{ p_1, ..., p_{m - 1} \} \\
& \boldsymbol{t} = \{ t_1, ..., t_l \} \\
\end{split}
\]
where~$m = \sum_i^n \mathbb{I}(r_i > 0)$ and~$l = \sum_i^{m - 1} p_i$. We define the policy as
\[
\begin{split}
    \pi(\boldsymbol{a} | \boldsymbol{y}) =
    & \prod_{r_i \in \boldsymbol{r}} \pi_{rps}(r_i \,|\, i, \boldsymbol{y}) \cdot \prod_{p_i \in \boldsymbol{p}} \pi_{plh}(p_i \,|\, i, \boldsymbol{y'}) \cdot \\
    & \prod_{t_i \in \boldsymbol{t}} \pi_{tok}(t_i \,|\, i, \boldsymbol{y''})
\end{split}
\]
with intermediate outputs~$\boldsymbol{y'} = \mathcal{E}(\boldsymbol{y}, \boldsymbol{r})$ and~$\boldsymbol{y''} = \mathcal{E}(\boldsymbol{y'}, \boldsymbol{p})$.

\begin{figure*}[!t]
    \centering
    \includegraphics[width=0.9\textwidth]{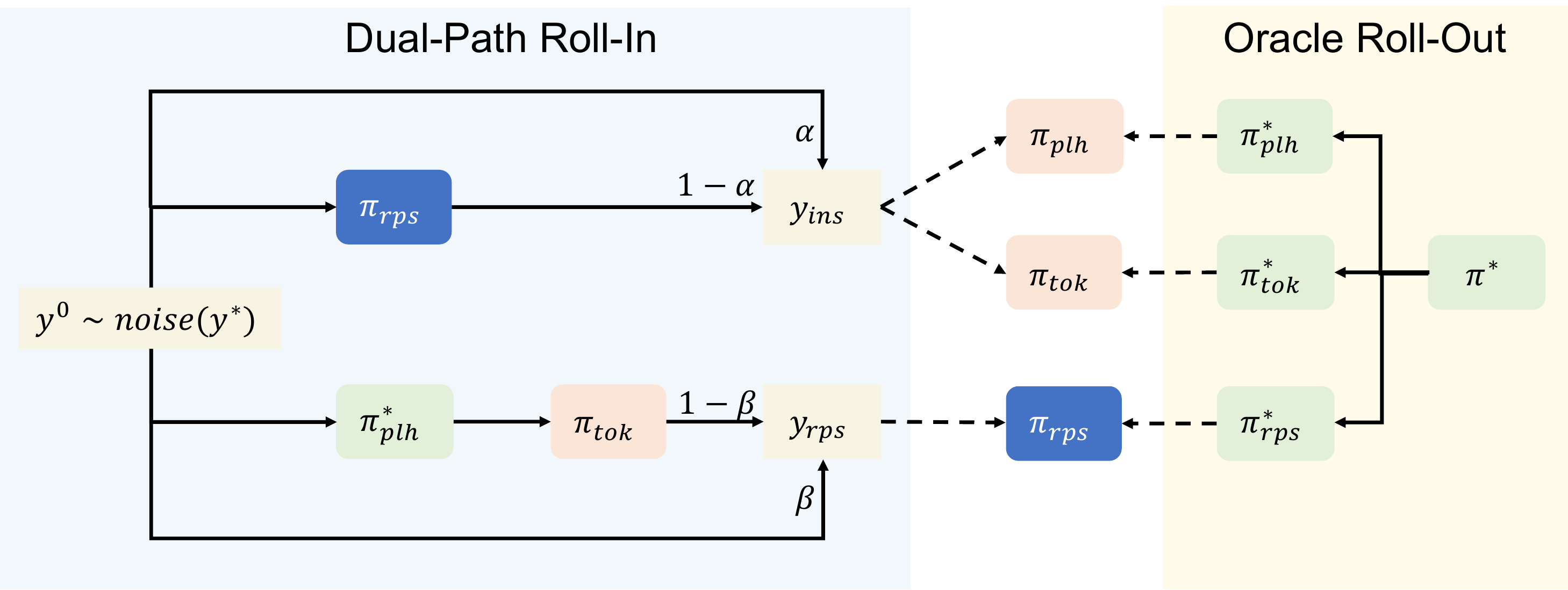}
\caption{Our dual-path imitation learning process uses both the reposition and insertion policies during roll-in so that they can be trained to refine each other's outputs: Given an initial sequence $\boldsymbol{y}^0$, created by noising the reference~$\boldsymbol{y}^*$, the roll-in policy stochastically generates intermediate sequences $\boldsymbol{y}_{ins}$ and $\boldsymbol{y}_{rps}$ via reposition and insertion respectively. The policy predictors are trained to minimize the costs of reaching $\boldsymbol{y}^*$ from $\boldsymbol{y}_{ins}$ and $\boldsymbol{y}_{rps}$ estimated by the oracle policy $\pi^*$.}
\label{fig:dual_path}
\end{figure*}

\subsection{Dual-Path Imitation Learning}

We train \modelname using imitation learning~\citep{DaumeLM2009,RossGB11,RossB2014} to efficiently explore the space of valid action sequences that can reach a reference translation. 
The key idea is to construct a \textit{roll-in} policy~$\pi^{in}$ to generate sequences to be refined and a \textit{roll-out} policy~$\pi^{out}$ to estimate cost-to-go for all possible actions given each input sequence. The model is trained to choose actions that minimizes the cost-to-go estimates. We use a search-based oracle policy~$\pi^*$ as the roll-out policy and train the model to imitate the optimal actions chosen by the oracle. 

Formally, $\boldsymbol{d}_{\pi_{rps}^{in}}$ and~$\boldsymbol{d}_{\pi_{ins}^{in}}$ denote the distributions of sequences induced by running the roll-in policies~$\pi_{rps}^{in}$ and~$\pi_{ins}^{in}$ respectively. We update the model policy~$\pi = \pi_{rps} \cdot \pi_{plh} \cdot \pi_{tok}$ to minimize the expected cost~$\mathcal{C}(\pi \,;\, \boldsymbol{y}, \pi^*)$ by comparing the model policy against the cost-to-go estimates under the oracle policy~$\pi^*$ given input sequences~$\boldsymbol{y}$:
\begin{equation}
\begin{split}
& \e{\boldsymbol{y}_{rps} \sim \boldsymbol{d}_{\pi_{rps}^{in}}}{\mathcal{C}(\pi_{rps} \,;\, \boldsymbol{y}_{rps}, \pi^*)} + \\
& \e{\boldsymbol{y}_{ins} \sim \boldsymbol{d}_{\pi_{ins}^{in}}}{\mathcal{C}(\pi_{plh}, \pi_{tok} \,;\, \boldsymbol{y}_{ins}, \pi^*)}
\end{split}
\end{equation}

The cost function compares the model vs.\ oracle actions. As prior work suggests that cost functions close to the cross-entropy loss are better suited to deep neural models  than the squared error \citep{LeblondAOL18,ChengYWB2018}, we define the cost function as the KL divergence between the action distributions given by the model policy and by the oracle~\citep{WelleckBDC2019}:
\begin{equation}
\begin{split}
& \mathcal{C}(\pi \,;\, \boldsymbol{y}, \pi^*) \\
=& \kld{\pi^*(\boldsymbol{a} \,|\, \boldsymbol{y}, \boldsymbol{y}^*)}{\pi(\boldsymbol{a} \,|\, \boldsymbol{y})} \\
=& \e{\boldsymbol{a} \sim \pi^*(\boldsymbol{a} \,|\, \boldsymbol{y}, \boldsymbol{y}^*)}{-\log \pi(\boldsymbol{a} \,|\, \boldsymbol{y})} + const. \\
\end{split}
\end{equation}
where the oracle has additional access to the reference sequence~$\boldsymbol{y}^*$. By minimizing the cost function, the model learns to imitate the oracle policy without access to the reference sequence.

Next, we describe how the reposition operation is incorporated in the roll-in policy~(Section~\ref{sec:roll_in}) and  the oracle roll-out policy~(Section~\ref{sec:oracle}).

\subsubsection{Dual-Path Roll-in Policy}
\label{sec:roll_in}

\begin{figure}[ht]
    \centering
    \includegraphics[width=0.5\textwidth]{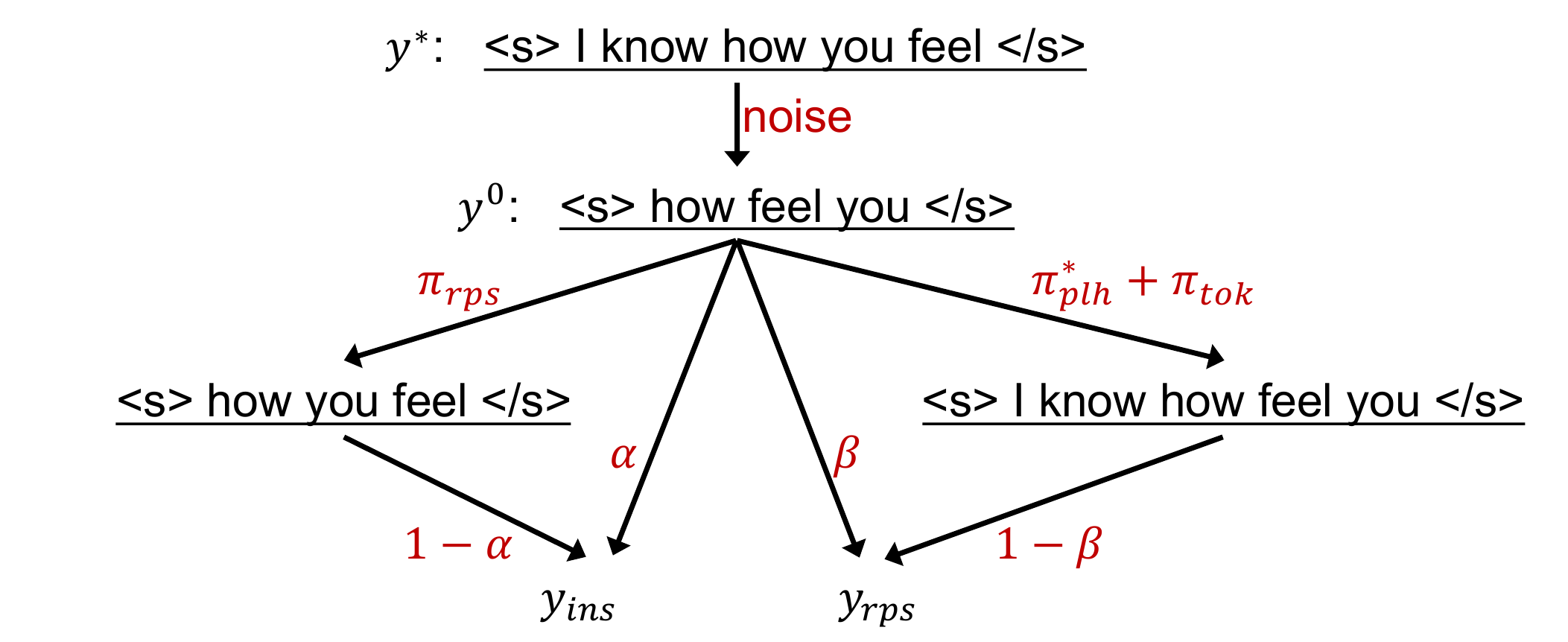}
\caption{The roll-in sequence for the insertion predictor is a stochastic mixture of the noised reference~$\boldsymbol{y}^0$ and the output by applying the model's reposition policy~$\pi_{rps}$ to~$\boldsymbol{y}^0$. The roll-in sequence for the reposition predictor is a stochastic mixture of the noised reference~$\boldsymbol{y}^0$ and the output by applying the oracle placeholder insertion policy~$\pi_{plh}^*$ and the model's token prediction policy~$\pi_{tok}$ to~$\boldsymbol{y}^0$.}
\label{fig:roll_in_example}
\end{figure}

As shown in Figure~\ref{fig:dual_path}, the roll-in policies~$\pi_{ins}^{in}$ and~$\pi_{rps}^{in}$ for the reposition and insertion policy predictors are stochastic mixtures of the noised reference sequences and the output sequences sampled from their corresponding dual policy predictors. Figure~\ref{fig:roll_in_example} shows an example for creating the roll-in sequences: we first create the initial sequence~$\boldsymbol{y}^0$ by applying random word dropping~\citep{GuWZ2019} and random word shuffle~\citep{LampleCDR18} with probability of~$0.5$ and maximum shuffle distance of~$3$ to the reference sequence~$\boldsymbol{y}^*$, and produce the roll-in sequences for each policy predictor as follows:
\begin{enumerate}
    \item \textbf{Reposition:} the roll-in policy~$\pi_{rps}^{in}$ is a stochastic mixture of the initial sequence~$\boldsymbol{y}^0$ and the output sequence by applying one iteration of the oracle placeholder insertion policy~$\boldsymbol{p}^* \sim \pi^*$ and the model's token prediction policy~$\tilde{\boldsymbol{t}} \sim \pi_{tok}$ to~$\boldsymbol{y}^0$:
        \begin{equation}
            \boldsymbol{d}_{\pi_{rps}^{in}} =
            \begin{cases}
                \boldsymbol{y}^0, & \text{if } u < \beta \\
                \mathcal{E}(\mathcal{E}(\boldsymbol{y}^0, \boldsymbol{p}^*), \tilde{\boldsymbol{t}}), & \text{otherwise}
            \end{cases}
        \end{equation}
    where the mixture factor~$\beta \in [0, 1]$ and random variable~$u \sim \text{Uniform}(0, 1)$.
    
    \item \textbf{Insertion:} the roll-in policy~$\pi_{ins}^{in}$ is a stochastic mixture of the initial sequence~$\boldsymbol{y}^0$ and the output sequence by applying one iteration of the model's reposition policy~$\tilde{\boldsymbol{r}} \sim \pi_{rps}$ to~$\boldsymbol{y}^0$:
        \begin{equation}
            \boldsymbol{d}_{\pi_{ins}^{in}} =
            \begin{cases}
                \boldsymbol{y}^0, & \text{if } u < \alpha \\
                \mathcal{E}(\boldsymbol{y}^0, \tilde{\boldsymbol{r}}), & \text{otherwise}
            \end{cases}
        \end{equation}
    where the mixture factor~$\alpha \in [0, 1]$ and random variable~$u \sim \text{Uniform}(0, 1)$.
\end{enumerate}

While \citet{GuWZ2019} define roll-in using only the model's insertion policy, we call our approach dual-path because roll-in creates two distinct intermediate sequences using the model's reposition or insertion policy. This makes it possible for the reposition and insertion policy predictors to learn to refine one another's outputs during roll-out, mimicking the iterative refinement process used at inference time.\footnote{Different from the inference process, we generate the roll-in sequences by applying the model's reposition or insertion policy for only one iteration.}

\subsubsection{Oracle Roll-Out Policy}
\label{sec:oracle}

\paragraph{Policy} Given an input sequence~$\boldsymbol{y}$ and a reference sequence~$\boldsymbol{y}^*$, the oracle algorithm finds the optimal action to transform~$\boldsymbol{y}$ into~$\boldsymbol{y}^*$ with the minimum number of basic edit operations:
\begin{equation}
\text{Oracle}(\boldsymbol{y}, \boldsymbol{y}^*) = \arg\min_{\boldsymbol{a}} \text{NumOps}(\boldsymbol{y}, \boldsymbol{y}^* \,|\, \boldsymbol{a})
\end{equation}
The associated oracle policy is defined as:
\begin{equation}
\pi^*(\boldsymbol{a} \,|\, \boldsymbol{y}, \boldsymbol{y}^*) =
\begin{cases}
    1, & \text{if } \boldsymbol{a} = \text{Oracle}(\boldsymbol{y}, \boldsymbol{y}^*) \\
    0, & \text{otherwise} \\
\end{cases}
\end{equation}

\paragraph{Algorithm} The reposition and insertion operations used in \modelname are designed so that the Levenshtein edit distance algorithm \citep{Levenshtein1966} can be used as the oracle. The reposition operation (Section~\ref{sec:basic_operations}) can be split into two distinct types of operations: (1) deletion and (2) replacing a word with any other word appearing in the input sequence, which is a constrained version of the Levenshtein substitution operation. As a result, we can use dynamic programming to find the optimal action sequence in~$O(|\boldsymbol{y}||\boldsymbol{y}^*|)$ time. By contrast, the Levenshtein Transformer restricts the oracle and model to insertion and deletion operations only. While in principle substitutions can be performed indirectly by deletion and re-insertion, our results show the benefits of using the reposition variant of the substitution operation. 

\subsection{Inference}
\label{sec:approach_inference}
During inference, we start from the initial sequence~$\boldsymbol{y}^0$. For standard sequence generation tasks,~$\boldsymbol{y}^0$ is an empty sequence, whereas for lexically constrained generation~$\boldsymbol{y}^0$ is a sequence of lexical constraints. 
Inference then proceeds in the exact same way for constrained and unconstrained tasks. The initial sequence is refined iteratively by applying a sequence of actions~$(\boldsymbol{a}^1, \boldsymbol{a}^2, ...) = (\boldsymbol{r}^1, \boldsymbol{p}^1, \boldsymbol{t}^1\,;\,\boldsymbol{r}^2, \boldsymbol{p}^2, \boldsymbol{t}^2\,;\,...)$.
We greedily select the best action at each iteration given the model policy in~\cref{eq:rps_classifier,eq:plh_classifier,eq:tok_classifier}. 
We stop refining if 
\begin{inparaenum}[1)]
\item the output sequences from two consecutive iterations are the same~\citep{GuWZ2019}, or
\item the maximum number of decoding steps is reached~\citep{LeeMC2018,GhazvininejadLLZ2019}.\footnote{Following~\citet{SternCKU2019}, we also experiment with adding penalty for inserting ``empty'' placeholders during inference by subtracting a penalty score~$\gamma = [0, 3]$ from the logits of zero in~\cref{eq:plh_classifier} to avoid overly short outputs. However, preliminary experiments show that zero penalty score achieves the best performance.}
\end{inparaenum}

\paragraph{Incorporating Soft Constraints}
Although \modelname is trained without lexical constraints, it can be used seamlessly for MT with constraints without any change to the decoding process except using the constraint sequence as the initial sequence.

\paragraph{Incorporating Hard Constraints}
We adopt the decoding technique introduced by \citet{SusantoCT2020} to enforce hard constraints at decoding time by prohibiting deletion operations on constraint tokens or insertions within a multi-token constraints.
\begin{table}
\centering
\begin{tabular}{lrrrr}
\toprule
 & Train & Valid & Test & Provenance \\\hline 
{Ro-En} & 599k & 1911 & 1999  & WMT16 \\ 
{En-De} & 3,961k & 3000 & 3003 & WMT14 \\ 
{En-Ja} & 2,000k & 1790 & 1812 & WAT2017 \\\hline
\end{tabular}
\caption{MT Tasks: data statistics (\# sentence pairs) and provenance per language pair.}
\label{tab:statistics}
\end{table}

\section{Experiments}
We evaluate the \modelname model on standard~(Section~\ref{sec:mt_exp}) and lexically constrained machine translation~(Sections~\ref{sec:lcmt_exp}--\ref{sec:exp_term}).

\subsection{Experimental Settings}

\paragraph{Dataset} 
Following~\citet{GuWZ2019}, we experiment on three language pairs spanning different language families and data conditions (Table~\ref{tab:statistics}): Romanian-English~(Ro-En) from WMT16~\citep{BojarWMT2016}, English-German~(En-De) from WMT14~\citep{BojarWMT2014}, and English-Japanese~(En-Ja) from WAT2017 Small-NMT Task~\citep{NakazawaWAT2017}. We also evaluate \modelname on the two En-De test sets with terminology constraints released by \citet{DinuMFA2019}. The test sets are subsets of the WMT17 En-De test set~\citep{Bojar2017WMT} with terminology constraints extracted from Wiktionary and IATE.\footnote{Available at
\url{https://www.wiktionary.org/} and \url{https://iate.europa.eu}.} For each test set, they only select the sentence pairs in which the exact target terms are used in the reference. The resulting Wiktionary and IATE test sets contain~727 and~414 sentences respectively. We follow the same preprocessing steps in~\citet{GuWZ2019}: we apply normalization, tokenization, true-casing, and BPE~\citep{SennrichHB16bpe} with 37k and 40k operations for En-De and Ro-En. For En-Ja, we use the provided subword vocabularies (16,384 BPE per language from SentencePiece~\citep{KudoR2018}).

\begin{table*}[ht]
\centering
\begin{tabular}{llccrrrr}
\toprule
& & \textbf{Distill} & \textbf{Beam} & {\bf Params} & \textbf{BLEU $\uparrow$} & \textbf{RIBES $\uparrow$} & \textbf{Latency~(ms) $\downarrow$} \\\hline
\multirow{5}{*}{Ro-En} & {\ar (fairseq)} & & 4 & 64.5M & 32.0 &	83.8 &	357.14 \\
& {\ar (sockeye)} & & 4 & 64.5M & 32.3 &	83.6 &	369.82 \\
& {\ar (sockeye)} & & 10 & 64.5M & 32.5 &	83.8 &	394.52 \\
& {\ar (sockeye)} & \checkmark & 10 & 64.5M & \underline{32.9} &	\underline{84.2} &	371.75 \\
& {\nar: \levt} & \checkmark & {--} & 90.9M & {\bf 31.6} &	{\bf 84.0} &	98.81 \\
& {\nar: \modelname} & \checkmark & {--} & 90.9M & {\bf 31.9} &	{\bf 84.0} &	\underline{\bf 93.20} \\
\hline
\multirow{5}{*}{En-De} & {\ar (fairseq)} && 4 & 64.9M & 27.1 &	80.4 &	363.64 \\
& {\ar (sockeye)} && 4 & 64.9M & \underline{27.3} &	80.2 &	308.64 \\
& {\ar (sockeye)} && 10 & 64.9M & \underline{27.4} &	80.3 &	332.73 \\
& {\ar (sockeye)} & \checkmark & 10 & 64.9M & \underline{27.6} &	80.5 &	363.52 \\
& {\nar: \levt} & \checkmark & {--} & 91.1M & {\bf 26.9} &	\underline{\bf 81.0} &	113.12 \\
& {\nar: \modelname} & \checkmark & {--} & 91.1M & {\bf 26.9} &	\underline{\bf 80.9} &	\underline{\bf 105.37} \\
\hline
\multirow{5}{*}{En-Ja} & {\ar (fairseq)} && 4 & 62.4M & \underline{44.9} &	\underline{85.7} &	292.40 \\
& {\ar (sockeye)} && 4 & 62.4M & 43.4 &	85.1 &	286.83 \\
& {\ar (sockeye)} && 10 & 62.4M & 43.5 &	85.3 &	311.38 \\
& {\ar (sockeye)} & \checkmark & 10 & 62.4M & 42.7 &	85.1 &	295.32 \\
& {\nar: \levt} & \checkmark & {--} & 106.1M & {\bf 42.4} &	84.5 &	143.88 \\
& {\nar: \modelname} & \checkmark & {--} & 106.1M & {\bf 42.3} &	{\bf 85.1} &	\underline{\bf 96.62} \\
\hline
\end{tabular}
\caption{Machine Translation Results. For each metric, we underline the top scores among all models and boldface the top scores among \nar models based on the paired bootstrap test with~$p < 0.05$ \citep{Clark2011}. \modelname decodes~6--7\% faster than \levt on Ro-En and En-De, and~33\% faster on En-Ja, while achieving comparable or higher BLEU and RIBES.}
\label{tab:mt}
\end{table*}

\paragraph{Experimental Conditions}
We train and evaluate the following models in controlled conditions to thoroughly evaluate \modelname:
\begin{itemize}
    \item \textbf{Auto-Regressive Transformers (\ar)} built using Sockeye~\citep{sockeye2017} and fairseq~\citep{Ott2019fairseq}. We report \ar baselines with both toolkits to enable fair comparisons when using our fairseq-based implementation of \modelname and Sockeye-based implementation of lexically constrained decoding algorithms \citep{PostV2018}.
    \item \textbf{Non Auto-Regressive Transformers (\nar)} In addition to \textbf{\modelname}, we train a Levenshtein Transformer (\textbf{\levt}) with approximately the same number of parameters. Both are implemented using fairseq.
\end{itemize}

\paragraph{Model \& Training Configurations}
All models adopt the \emph{base} Transformer architecture~\citep{Vaswani2017} with~$d_{\text{model}}=512$,~$d_{\text{hidden}}=2048$,~$n_{\text{heads}}=8$,~$n_{\text{layers}}=6$, and~$p_{\text{dropout}} = 0.3$. For En-De and Ro-En, the source and target embeddings are tied with the output layer weights~\citep{PressW17,NguyenC18}.
We add dropout to embeddings~(0.1) and label smoothing~(0.1). \ar models are trained with the Adam optimizer~\citep{KingmaB15} with a batch size of~4096 tokens. We checkpoint models every~1000 updates.  The initial learning rate is~0.0002, and it is reduced by~30\% after~4 checkpoints without validation perplexity improvement. Training stops after~20 checkpoints without improvement. All \nar models are trained using Adam ~\citep{KingmaB15} with initial learning rate of~0.0005 and a batch size of 64,800 tokens for maximum 300,000 steps.\footnote{Our preliminary experiments and prior work show that \nar models require larger training batches than \ar models.} We select the best checkpoint based on validation BLEU~\citep{Papineni2002BLEU}. All models are trained on 8 NVIDIA V100 Tensor Core GPUs.

\paragraph{Knowledge Distillation}
We apply sequence-level knowledge distillation from autoregressive teacher models as widely used in non-autoregressive generation~\citep{GuBXLS2018,LeeMC2018,GuWZ2019}.  Specifically, when training the non-autoregressive models, we replace the reference sequences~$\boldsymbol{y}^*$ in the training data with translation outputs from the \ar teacher model~(Sockeye, with~$\text{beam}=4$).\footnote{This teacher model was selected for a fairer comparison on MT with lexical constraints.}
We also report the results when applying knowledge distillation to autoregressive models.

\paragraph{Evaluation}
We evaluate translation quality via case-sensitive tokenized \textbf{BLEU} (as in ~\citet{GuWZ2019})\footnote{\url{https://github.com/pytorch/fairseq/blob/master/fairseq/clib/libbleu/libbleu.cpp}}
and \textbf{RIBES}~\citep{Isozaki2010RIBES}, which is more sensitive to word order differences. Before computing the scores, we tokenize the German and English outputs using Moses and Japanese outputs using KyTea.\footnote{\url{http://www.phontron.com/kytea/}} For lexically constrained decoding, we report the constraint preservation rate (\textbf{CPR}) in the translation outputs.

We quantify decoding speed using \textbf{latency} per sentence. It is computed as the average time (in ms) required to translate the test set using batch size of one~(excluding the model loading time) divided by the number of sentences in the test set.

\subsection{MT Tasks}
\label{sec:mt_exp}

Since our experiments involve two different toolkits, we first compare the same Transformer \ar models built with Sockeye and with fairseq: the \ar models achieve comparable decoding speed and translation quality regardless of toolkit \---\ the Sockeye model obtains higher BLEU than the fairseq model on Ro-En and En-De but lower on En-Ja (Table~\ref{tab:mt}).  Further comparisons will therefore center on the Sockeye \ar model to better compare \modelname with the lexically constrained decoding algorithm~\citep{PostV2018}.

Table~\ref{tab:mt} also shows that knowledge distillation has a small and inconsistent impact on \ar models (Sockeye): it yields higher BLEU on Ro-En, close BLEU on En-De, and lower BLEU on En-Ja.\footnote{\citet{KasaiPPCS2020} found that \ar models can benefit from knowledge distillation but with a Transformer large model as a teacher, while we use the Transformer base model.} Thus, we use the \ar models trained without distillation in further experiments.

Next, we compare the \nar models against the \ar~(Sockeye) baseline. As expected, both \modelname and \levt achieve close translation quality to their \ar teachers with~2--4 times speedup. BLEU differences are small~($\Delta < 1.1$) as in prior work \citep{GuWZ2019}. The RIBES trends are more surprising: both \nar models significantly outperform the \ar models~(Sockeye) on RIBES, except for En-Ja, where \modelname and the \ar models significantly outperforms \levt. This illustrates the strength of \modelname in word reordering.

Finally, results confirm the benefits of \modelname's reposition operation over \levt: decoding with \modelname is 6--7\% faster than \levt on Ro-En and En-De, and~33\% faster on En-Ja \---\ a more distant language pair which requires more reordering but no inflection changes on reordered words \---\ with no statistically significant difference in BLEU nor RIBES, except for En-Ja, where \modelname significantly outperforms \levt on RIBES.
Overall, \modelname is shown to be a good alternative to \levt on standard machine translation tasks and can also be used to replace the \ar models in settings where decoding speed matters more than small differences in translation quality.

\begin{table*}[ht]
\centering
\begin{tabular}{llccrrrr}
\toprule
& & \textbf{Distill} & \textbf{Beam} & \textbf{BLEU $\uparrow$} & \textbf{RIBES $\uparrow$} & \textbf{CPR $\uparrow$} & \textbf{Latency~(ms) $\downarrow$} \\\hline
\multirow{4}{*}{Ro-En} & {\ar + DBA (sockeye)} && 4 & 31.0 &	79.5 &	\underline{99.7} &	436.26 \\
& {\ar + DBA (sockeye)} && 10 & \underline{34.6} &	84.5 &	99.5 &	696.68 \\
& {\nar: \levt} & \checkmark & {--} & 31.6 &	83.4 &	80.3 &	121.80 \\
& {+ hard constraints} & \checkmark & {--} & 27.7	& 78.4	& 99.9	& 140.79 \\
& {\nar: \modelname} & \checkmark & {--} & {\bf 33.1} &	\underline{\bf 85.0} &	{\bf 86.8} &	\underline{\bf 108.98} \\
& {+ hard constraints} & \checkmark & {--} & 28.8	& 81.2	& 95.0	& 136.78 \\
\hline
\multirow{4}{*}{En-De} & {\ar + DBA (sockeye)} && 4 & 26.1 &	74.7 &   \underline{99.7} &	434.41 \\
& {\ar + DBA (sockeye)} && 10 & \underline{30.5} &	\underline{81.9} &	99.5 &	896.60 \\
& {\nar: \levt} & \checkmark & {--} & 27.1 &	80.0 &	75.6 &	127.00 \\
& {+ hard constraints} & \checkmark & {--} & 24.9 &	74.1 &	100.0 &	134.10 \\
& {\nar: \modelname} & \checkmark & {--} & {\bf 28.2} &	{\bf 81.6} &	{\bf 88.4} &	\underline{\bf 121.65} \\
& {+ hard constraints} & \checkmark & {--} & 25.8 &	77.2 &	96.8 &	134.10 \\
\hline
\multirow{4}{*}{En-Ja} & {\ar + DBA (sockeye)} && 4 & 44.3 &	81.6 &	\underline{100.0} &	418.71 \\
& {\ar + DBA (sockeye)} && 10 & \underline{48.0} &	\underline{85.9} &	\underline{100.0} &	736.92 \\
& {\nar: \levt} & \checkmark & {--} & 42.8 &	84.0 &	74.3 &	161.17 \\
& {+ hard constraints} & \checkmark & {--} & 39.7 &	77.4 &	99.9 &	159.27 \\
& {\nar: \modelname} & \checkmark & {--} & {\bf 45.3} &	{\bf 85.7} &	{\bf 91.3} &	\underline{\bf 109.50} \\
& {+ hard constraints} & \checkmark & {--} & 43.7 &	82.6 &	96.4 &	132.71 \\
\hline
\end{tabular}
\caption{Machine Translation with lexical constraints~(averages over~5 runs). For each metric, we underline the top scores among all models and boldface the top scores among \nar models based on the independent student's t-test with~$p < 0.05$. \modelname exploits constraints better than \levt. It also achieves comparable RIBES to the best \ar model with~6--7 times decoding speedup.}
\label{tab:lcmt}
\end{table*}

\subsection{MT with Lexical Constraints}
\label{sec:lcmt_exp}

We now turn to the main evaluation of \modelname on machine translation with lexical constraints.  

\paragraph{Experimental Conditions}
We conduct a controlled comparison of the following approaches:
\begin{itemize}
    \item \nar models: \textbf{\modelname} and \textbf{\levt} view the lexical constraints as \textbf{soft constraints}, provided via the initial target sequence. We also explore the decoding technique introduced in \citet{SusantoCT2020} to support \textbf{hard constraints}.
    \item \ar models: they use the provided target words as hard constraints enforced at decoding time by an efficient form of constrained beam search: dynamic beam allocation (\textbf{DBA})~\citep{PostV2018}.\footnote{Although the beam pruning option in \citet{PostV2018} is not used here (since it is not supported in Sockeye anymore), other Sockeye updates improve efficiency. Constrained decoding with DBA is~1.8--2.7 times slower than unconstrained decoding here, while DBA is~3 times slower when~$\text{beam}=10$ in \citet{PostV2018}.}
\end{itemize}
Crucially, all models, including \modelname, are the exact same models evaluated on the standard MT tasks above, and do not need to be trained specifically to incorporate constraints.

We define lexical constraints as \citet{PostV2018}: for each source sentence, we randomly select one to four words from the reference as lexical constraints. We then randomly shuffle the constraints and apply BPE to the constraint sequence. Different from the terminology test sets in \citet{DinuMFA2019} which contain only several hundred sentences with mostly nominal constraints, our constructed test sets are larger and include lexical constraints of all types.

\begin{figure}[!t]
    \centering
    \begin{subfigure}[b]{0.42\textwidth}
        \centering
        \includegraphics[width=\textwidth]{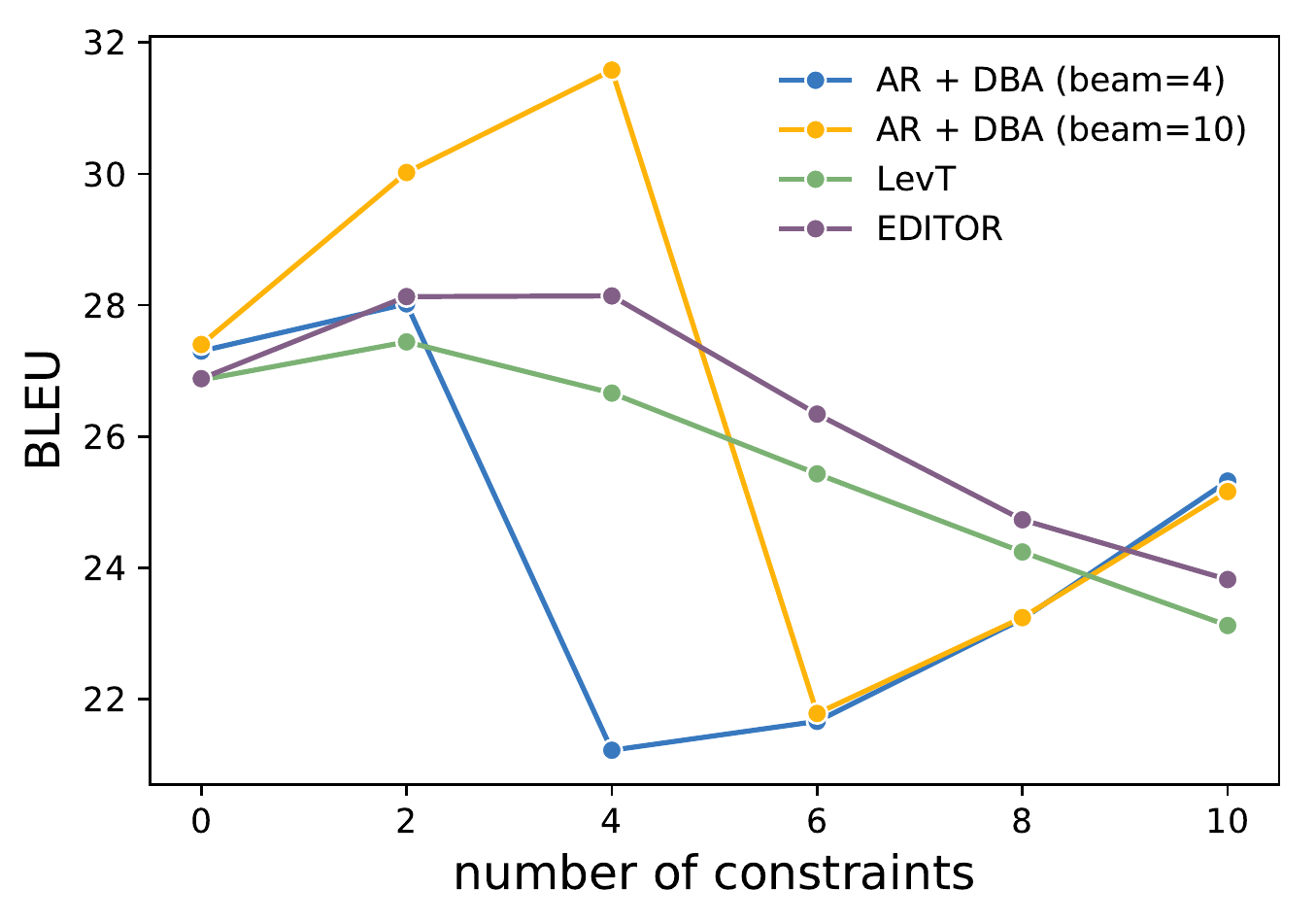}
        \caption{En-De}
    \end{subfigure}

	\centering
    \begin{subfigure}[b]{0.42\textwidth}
        \centering
        \includegraphics[width=\textwidth]{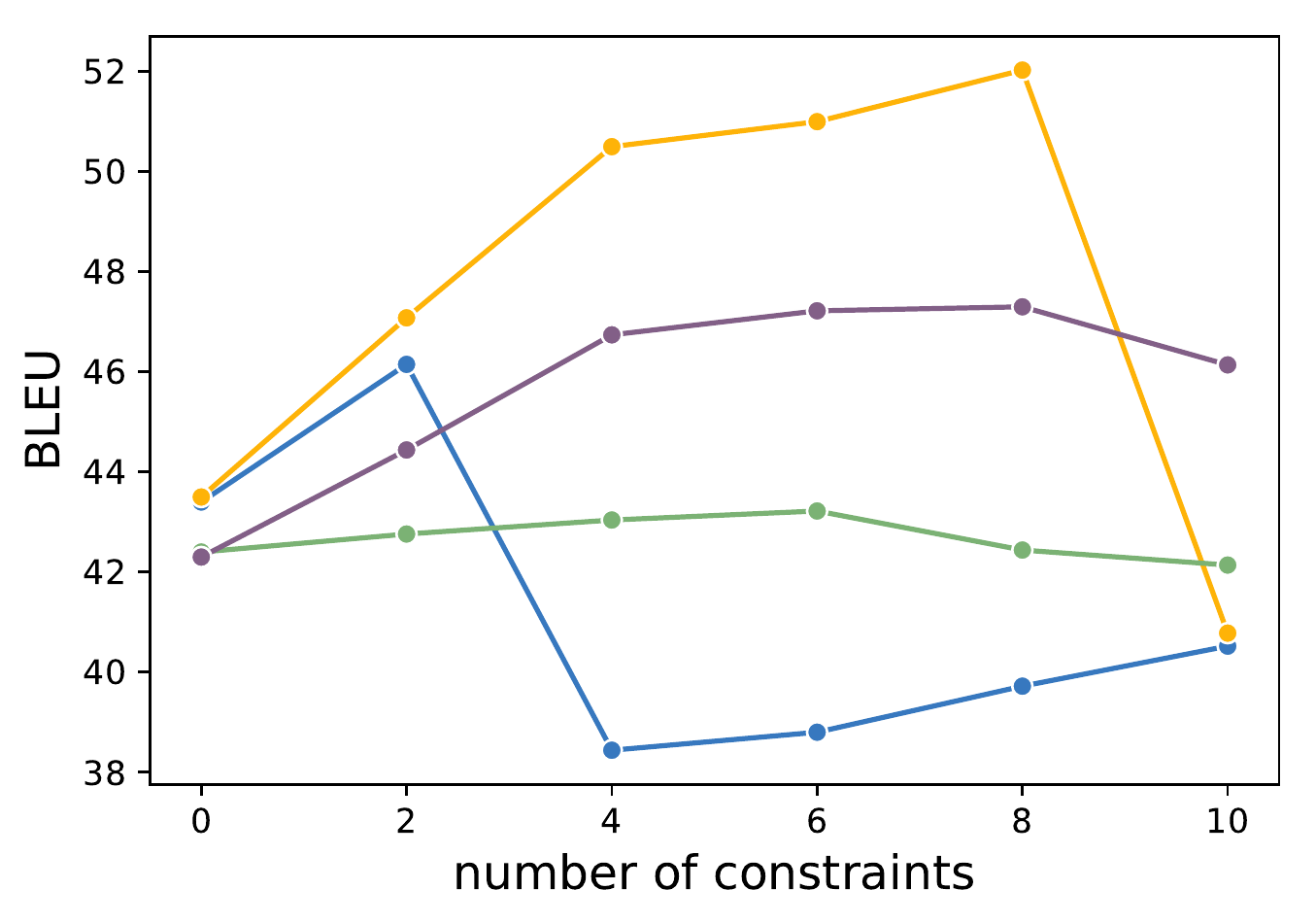}
        \caption{En-Ja}
    \end{subfigure}

	\centering
    \begin{subfigure}[b]{0.42\textwidth}
        \centering
        \includegraphics[width=\textwidth]{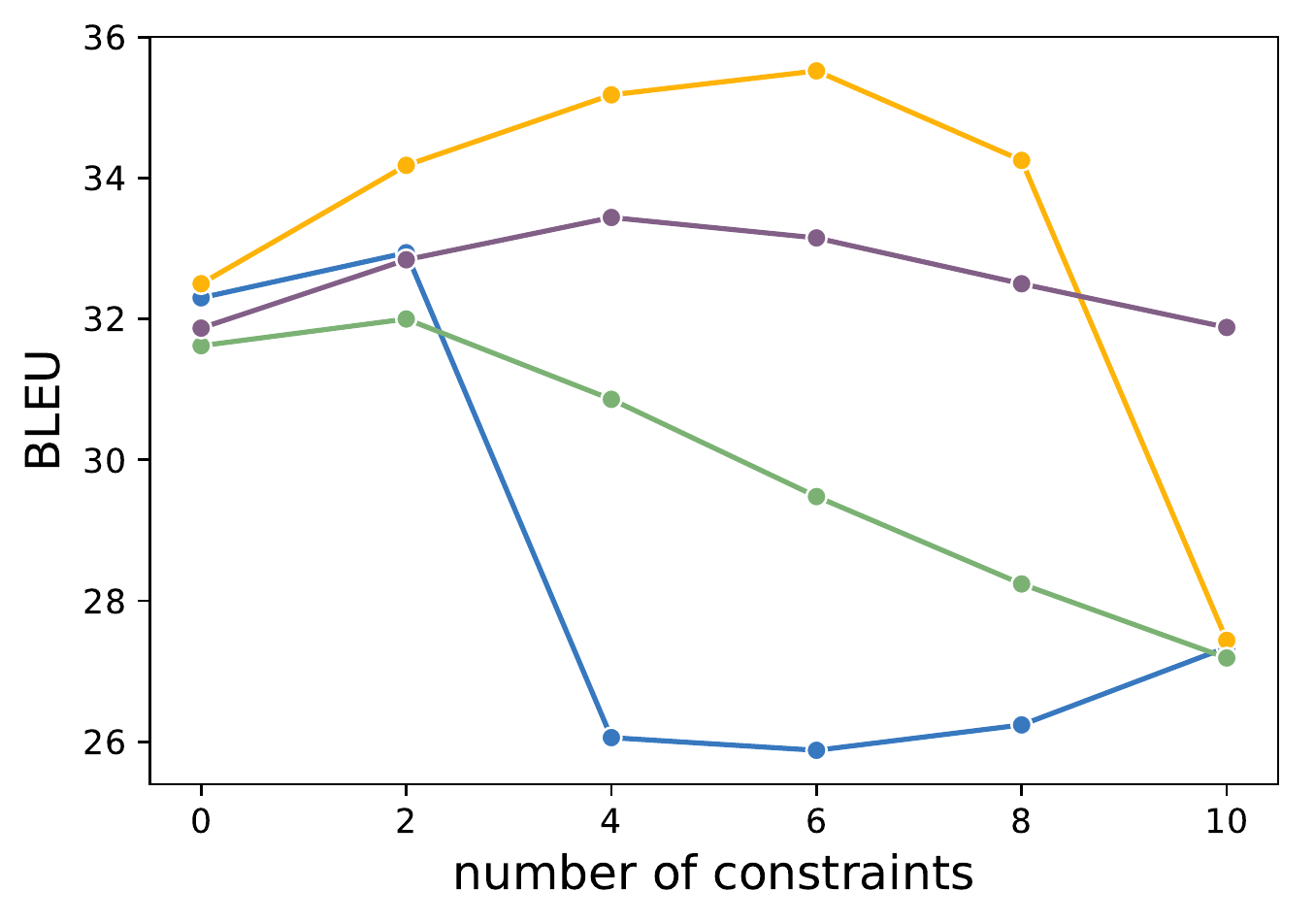}
        \caption{Ro-En}
    \end{subfigure}
\caption{\modelname improves BLEU over \levt for 2--10 constraints (counted pre-BPE) and beats the best \ar model on 2/3 tasks with 10 constraints.}
\label{fig:bleu_kconst}
\end{figure}

\begin{figure}[!t]
    \centering
    \begin{subfigure}[b]{0.42\textwidth}
        \centering
        \includegraphics[width=\textwidth]{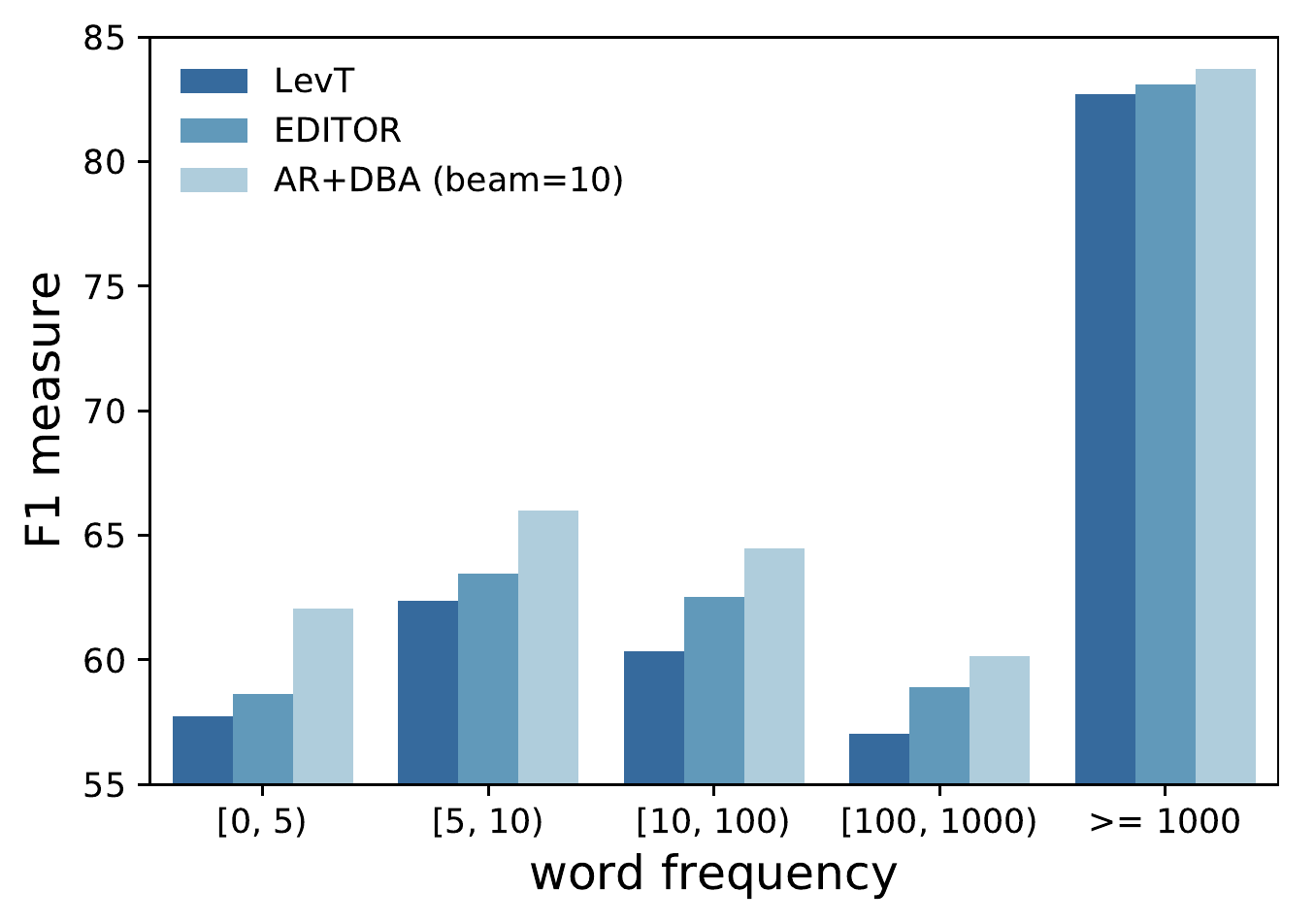}
        \caption{En-De}
    \end{subfigure}
	\hfill
	
	\centering
    \begin{subfigure}[b]{0.42\textwidth}
        \centering
        \includegraphics[width=\textwidth]{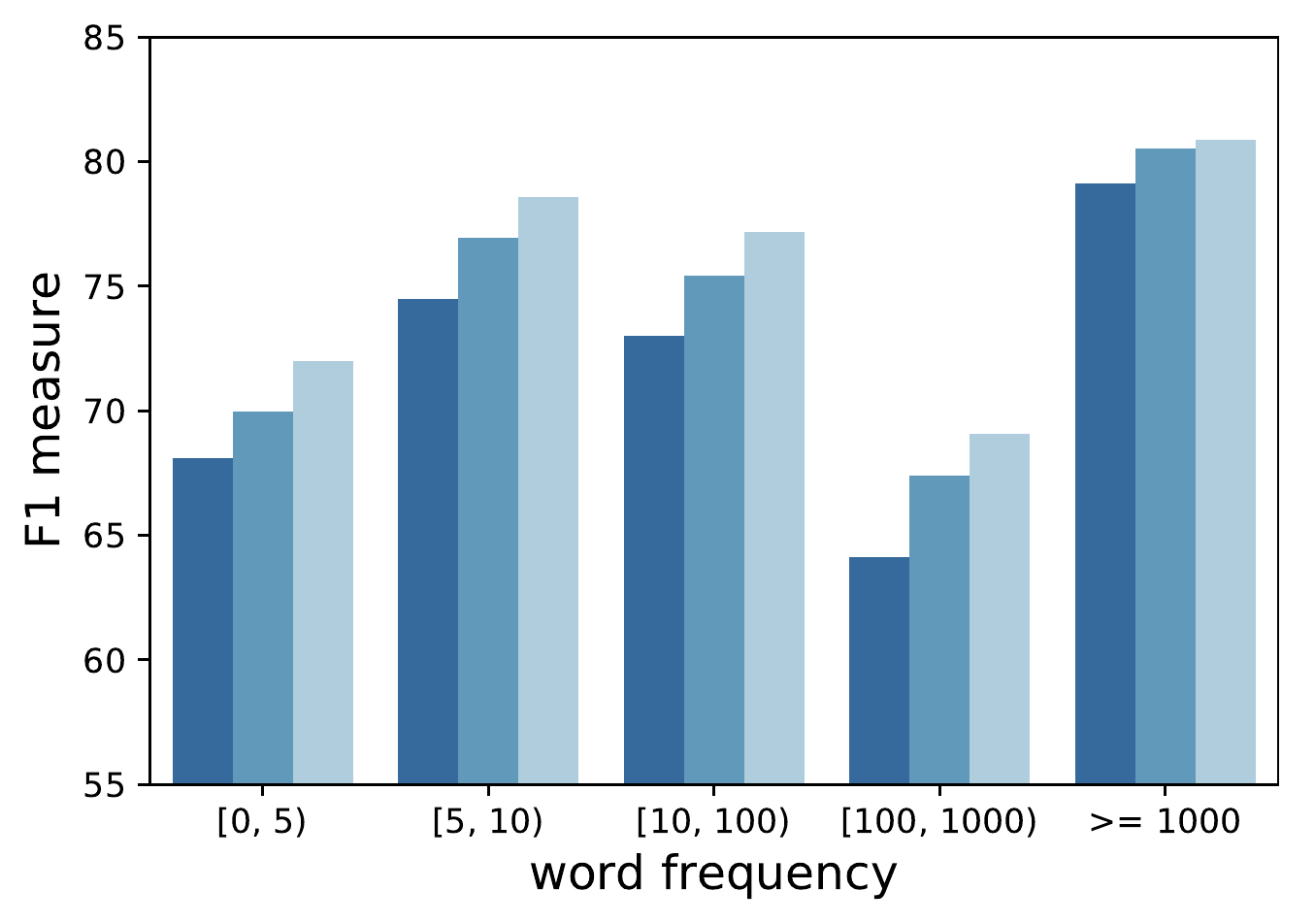}
        \caption{En-Ja}
    \end{subfigure}
	\hfill
	
	\centering
    \begin{subfigure}[b]{0.42\textwidth}
        \centering
        \includegraphics[width=\textwidth]{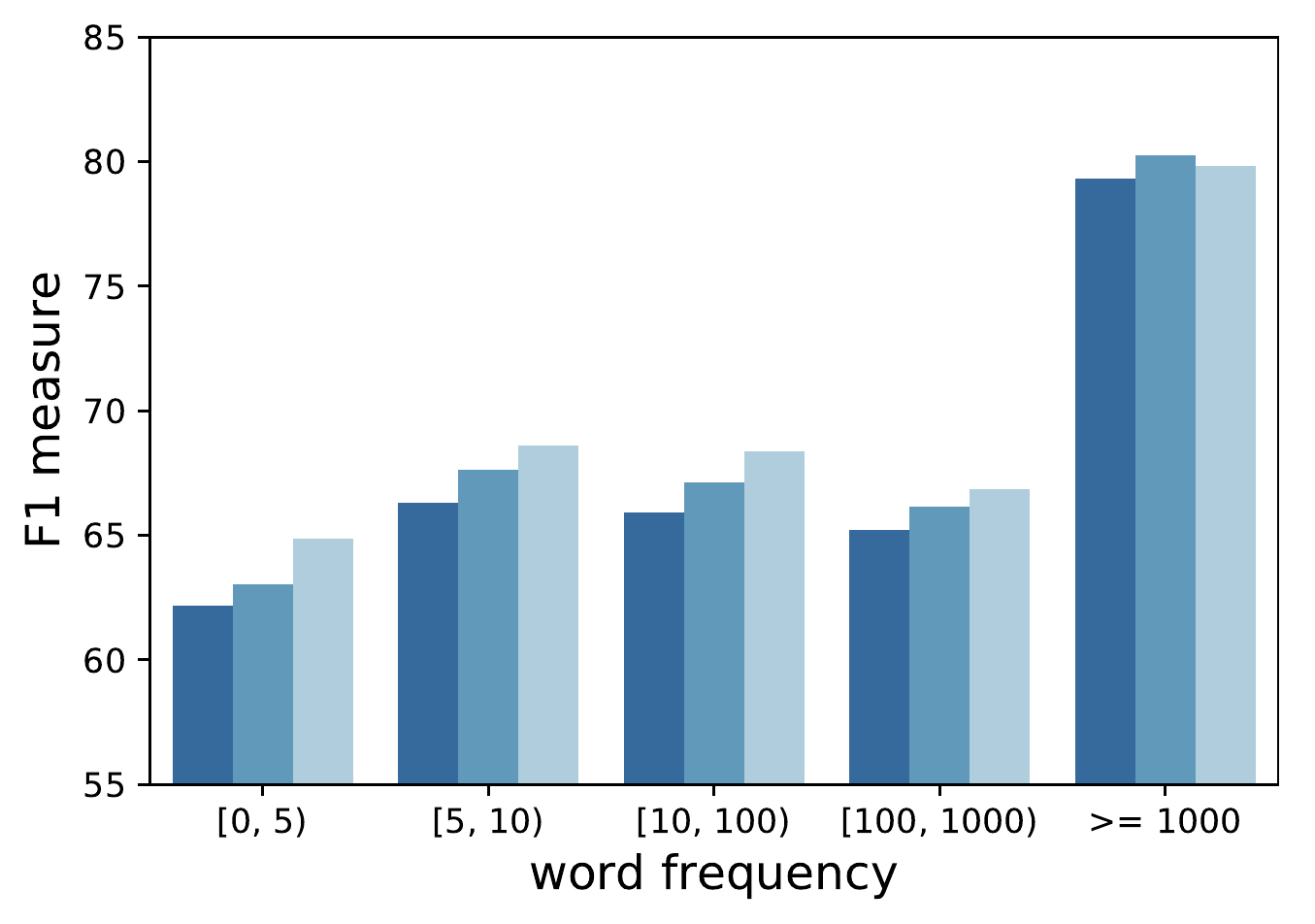}
        \caption{Ro-En}
    \end{subfigure}
\caption{Target word F1 score binned by word test set frequency: \modelname improves over \levt the most for  words of low or medium frequency. \ar achieves higher F1 than \modelname for words of low or medium frequency at the cost of much longer decoding time.}
\label{fig:acc_freq}
\end{figure}

\paragraph{Main Results}
Table~\ref{tab:lcmt} shows that \modelname exploits the soft constraints to strike a better balance between translation quality and decoding speed than other models. Compared to \levt, \modelname preserves~7--17\% more constraints and achieves significantly higher translation quality~(+1.1--2.5 on BLEU and +1.6--1.8 on RIBES) and faster decoding speed.
Compared to the \ar model with~$\text{beam}=4$, \modelname yields significantly higher BLEU~(+1.0--2.2) and RIBES~(+4.1--6.9) with~3--4 times decoding speedup.
After increasing the beam to $10$, \modelname obtains lower BLEU but comparable RIBES with~6--7 times decoding speedup.\footnote{\citet{PostV2018} show that the optimal beam size for DBA is~20. Our experiment on En-De shows that increasing the beam size from~10 to~20 improves BLEU by~0.7 at the cost of doubling the decoding time.} Note that \ar models treat provided words as hard constraints and therefore achieve over 99\% CPR by design, while \nar models treat them as soft constraints.

Results confirm that enforcing hard constraints increases CPR but degrades translation quality compared to the same model using soft constraints: for \levt, it degrades BLEU by~2.2--3.9 and RIBES by~5.0--6.6. For \modelname, it degrades BLEU by~1.6--4.3 and RIBES by~3.1--4.4~(Table~\ref{tab:lcmt}). By contrast, \modelname with soft constraints strikes a better balance between translation quality and constraint preservation.

The strengths of \modelname hold when varying the number of constraints (Figure~\ref{fig:bleu_kconst}). For all tasks and models, adding constraints helps BLEU up to a certain point, ranging from 4 to 10 words. When excluding the slower \ar model ($\text{beam} = 10$), \modelname consistently reaches the highest BLEU score with 2--10 constraints: \modelname outperforms \levt and the \ar model with~$\text{beam} = 4$.  
Consistent with ~\citet{PostV2018}, as the number of constraints increases, the \ar model needs larger beams to reach good performance. When the number of constraints increases to 10, \modelname yields higher BLEU than the \ar model on En-Ja and Ro-En, even after incurring the cost of increasing the \ar beam to $10$.

Are \modelname improvements limited to preserving constraints better? We verify that this is not the case by computing the target word F1 binned by frequency~\citep{Neubig2019compare}. 
Figure~\ref{fig:acc_freq} shows that \modelname improves over \levt across all test frequency classes and closes the gap between \nar and \ar models: the largest improvements are obtained for low and medium frequency words \---\ on En-De and En-Ja, the largest improvements are on words with frequency between~5 and~1000, while on Ro-En, \modelname improves more on words with frequency between~5 and~100. \modelname also improves F1 on rare words~(frequency in $[0, 5)$), but not as much as for more frequent words.

We now conduct further analysis to better understand the factors that contribute to \modelname's advantages over \levt.

\begin{table}[!t]
\centering
\begin{tabular}{lrrrrr}
\toprule
& Repos. & Del. &  Ins. & Total & Iter. \\\hline
\textit{Ro-En} \\
{\levt} & 0.00 &	4.61 &	33.05 &	37.67 & 2.01 \\
{\modelname} & 8.13 &	2.50 &	28.68 &	39.31 & 1.81 \\
\hline
\textit{En-De} \\
{\levt} & 0.00 &	7.13 &	45.45 &	52.58 & 2.14 \\
{\modelname} & 5.85 &	4.01 &	28.75 &	38.61 & 2.07 \\
\hline
\textit{En-Ja} \\
{\levt} & 0.00 &	5.24 &	32.83 &	38.07 & 2.93 \\
{\modelname} & 4.73 &	1.69 &	21.64 &	28.06 & 1.76 \\
\hline
\end{tabular}
\caption{Average number of repositions~(excluding deletions), deletions, insertions, and decoding iterations to translate each sentence with soft lexical constraints~(averaged over 5 runs). Thanks to reposition operations, \modelname uses~40--70\% fewer deletions,~10--40\% fewer insertions, and~3--40\% fewer decoding iterations overall.}
\label{tab:num_ops}
\end{table}

\begin{figure*}[!t]
    \centering
    \includegraphics[width=0.72\textwidth]{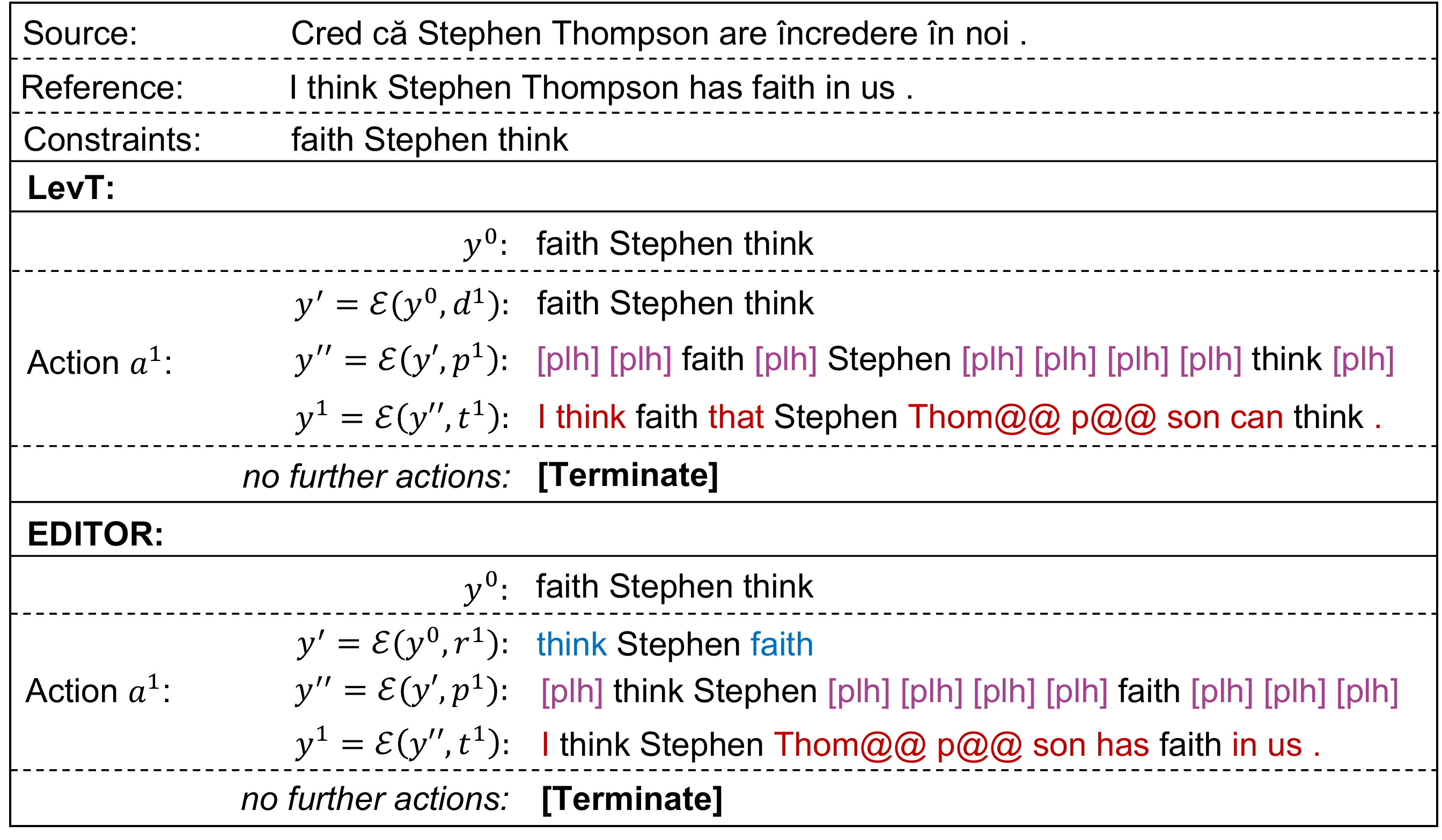}
\caption{Ro-En translation with soft lexical constraints:  while \levt uses the constraints in the provided order, \modelname's reposition operation helps generate a more fluent and adequate translation.}
\label{fig:example_roen}
\end{figure*}

\begin{table}[ht]
\centering
\begin{tabular}{lrrrr}
\toprule
& BLEU$\uparrow$ & RIBES$\uparrow$ & CPR$\uparrow$ & Lat.$\downarrow$ \\\hline
\textit{Ro-En} \\
{\modelname} & 33.1 &	85.0 &	86.8 &	108.98 \\
{-dual-path} & 32.2 &	84.4 &	74.8 &	119.61 \\
{\levt} & 31.6 &	83.4 &	80.3 &	121.80 \\
\hline
\textit{En-De} \\
{\modelname} & 28.2 &	81.6 &	88.4 &	121.65 \\
{-dual-path} & 27.2 &	80.4 &	78.7 &	130.85 \\
{\levt} & 27.1 &	80.0 &	75.6 &	127.00 \\
\hline
\textit{En-Ja} \\
{\modelname} & 45.3 &	85.7 &	91.3 &	109.50 \\
{-dual-path} & 44.0 &	83.9 &	80.0 &	154.10 \\
{\levt} & 42.8 &	84.0 &	74.3 &	161.17 \\
\hline
\end{tabular}
\caption{Ablating the dual-path roll-in policy hurts \modelname on soft-constrained MT, but still outperforms \levt, confirming that reposition and dual-path imitation learning both benefit \modelname.}
\label{tab:ablation}
\end{table}

\begin{table*}[!t]
\centering
\begin{tabular}{lrrrr}
\toprule
& \multicolumn{2}{c}{Wiktionary} & \multicolumn{2}{c}{IATE} \\
& Term\%$\uparrow$ & BLEU$\uparrow$ & Term\%$\uparrow$ & BLEU$\uparrow$ \\
\hline
\multicolumn{5}{l}{Prior Results} \\
%\hline
{Base Trans.} & 76.9 &	26.0 &	76.3 &	25.8 \\
{Post18} & 99.5 &	25.8 &	82.0 &	25.3 \\
{Dinu19} & 93.4 &	26.3 &	94.5 &	26.0 \\
{Base \levt} & 81.1 &	30.2 &	80.3 &	29.0 \\
{Susanto20} & 100.0 &	31.2 &	100.0 & 30.1 \\
\hline
\multicolumn{5}{l}{Our Results} \\
{\levt} & 84.3 &	28.2 &	83.9 &	27.9 \\
{+ soft constraints} & 90.5 &	28.5 &	92.5 &	28.3 \\
{+ hard constraints} & 100.0 &	28.8 &	100.0 & 28.9 \\
{\modelname} & 83.5 &	28.8 &	83.0 &	27.9 \\
{+ soft constraints} & 96.8 &	29.3 &	97.1 &	28.8 \\
{+ hard constraints} & 99.8 &	29.3 &	100.0 &	28.9 \\
\hline
\end{tabular}
\caption{Term usage percentage~(\textit{Term\%}) and BLEU scores of En-De models on terminology test sets~\citep{DinuMFA2019} provided with correct terminology entries (exact matches on both source and target sides). \modelname with soft constraints achieves higher BLEU than \levt with soft constraints, and on par or higher BLEU than \levt with hard constraints.}
\label{tab:term_exact}
\end{table*}

\paragraph{Impact of Reposition} We compare the average number of basic edit operations~(Section~\ref{sec:basic_operations}) of different types used by \modelname and \levt on each test sentence (averaged over the 5 runs): reposition~(excluding deletion for controlled comparison with \levt), deletion, and insertion performed by \levt and \modelname at decoding time. Table~\ref{tab:num_ops} shows that \levt deletes tokens~2--3 times more often than \modelname, which explains its lower CPR than \modelname. \levt also inserts tokens~1.2--1.6 times more often than \modelname and performs~1.4 times more edit operations on En-De and En-Ja. On Ro-En, \levt performs -4\% fewer edit operations in total than \modelname but is overall slower than \modelname, since multiple operations can be done in parallel at each action step. Overall, \modelname takes~3--40\% fewer decoding iterations than \levt. These results suggest that reposition successfully reduces redundancy in edit operations and makes decoding more efficient by replacing sequences of insertions and deletions with a single repositioning step.

Furthermore, Figure~\ref{fig:example_roen} illustrates how reposition increases flexibility in exploiting lexical constraints, even when they are provided in the wrong order. While \levt generates an incorrect output by using constraints in the provided order, \modelname's reposition operation helps generate a more fluent and adequate translation.

\paragraph{Impact of Dual-Path Roll-In} Ablation experiments (Table~\ref{tab:ablation}) show that \modelname benefits greatly from dual-path roll-in.
Replacing dual-path roll-in with the simpler roll-in policy used in~\citet{GuWZ2019}, the model's translation quality drops significantly~(by~0.9--1.3 on BLEU and~0.6--1.9 on RIBES) with fewer constraints preserved and slower decoding. It still achieves better translation quality than \levt thanks to the reposition operation: specifically, it yields significantly higher BLEU and RIBES on Ro-En, comparable BLEU and significantly higher RIBES on En-De, and comparable RIBES and significantly higher BLEU on En-Ja than \levt.

\subsection{MT with Terminology Constraints}
\label{sec:exp_term}
We evaluate \modelname on the terminology test sets released by \citet{DinuMFA2019} to test its ability to incorporate terminology constraints and to further compare it with prior work \citep{DinuMFA2019,PostV2018,SusantoCT2020}.

Compared to \citet{PostV2018} and \citet{DinuMFA2019}, \modelname with soft constraints achieves higher absolute BLEU, and higher BLEU improvements over its counterpart without constraints~(Table~\ref{tab:term_exact}). Consistent with previous findings by \citet{SusantoCT2020}, incorporating soft constraints in \levt improves BLEU by~+0.3 on Wiktionary and by~+0.4 on IATE. Enforcing hard constraints as in \citet{SusantoCT2020} increases the term usage by~+8--10\% and improves BLEU by~+0.3--0.6 over \levt using soft constraints.\footnote{We use our implementations of \citet{SusantoCT2020}'s technique for a more controlled comparison. The \levt baseline in \citet{SusantoCT2020} achieves higher BLEU than ours on the small Wiktionary and IATE test sets, while it underperforms our \levt on the full WMT14 test set (26.5 vs. 26.9).} For \modelname, adding soft constraints improves BLEU by~+0.5 on Wiktionary and~+0.9 on IATE, with very high term usages (96.8\% and 97.1\% respectively). \modelname thus correctly uses the provided terms almost all the time when they are provided as soft constraints, so there is little benefit to enforcing hard constraints instead: they help close the small gap to reach 100\% term usage and do not improve BLEU.
Overall, \modelname achieves on par or higher BLEU than \levt with hard constraints.

Results also suggest that \modelname can handle phrasal constraints even though it relies on token-level edit operations, since it achieves above~99\% term usage on the terminology test sets where~26--27\% of the constraints are multi-token.

\section{Conclusion}
We introduce \modelname, a non-autoregressive transformer model that iteratively edits hypotheses using a novel reposition operation.  Reposition combined with a new dual-path imitation learning strategy helps \modelname generate output sequences that flexibly incorporate user's lexical choice preferences.
Extensive experiments show that \modelname exploits soft lexical constraints more effectively than the Levenshtein Transformer~\citep{GuWZ2019} while speeding up decoding dramatically compared to constrained beam search~\citep{PostV2018}.
Results also confirm the benefits of using soft constraints over hard ones in terms of translation quality.
\modelname also achieves comparable or better translation quality with faster decoding speed than the Levenshtein Transformer on three standard MT tasks. These promising results open several avenues for future work, including using \modelname for other generation tasks than MT and investigating  its ability to incorporate more diverse constraint types into the decoding process. 

\section*{Acknowledgments}
We thank Sweta Agrawal, Kiant\'e Brantley, Eleftheria Briakou, Hal Daum\'e III, Aquia Richburg, Fran{\c{c}}ois Yvon, the TACL reviewers, and the CLIP lab at UMD for their helpful and constructive comments. This research is supported in part by an Amazon Web Services Machine Learning Research Award and by the Office of the Director of National Intelligence (ODNI), Intelligence Advanced Research Projects Activity (IARPA), via contract \#FA8650-17-C-9117. The views and conclusions contained herein are those of the authors and should not be interpreted as necessarily representing the official policies, either expressed or implied, of ODNI, IARPA, or the U.S. Government. The U.S. Government is authorized to reproduce and distribute reprints for governmental purposes notwithstanding any copyright annotation therein.

\bibliography{anthology,emnlp2020}

\begin{thebibliography}{60}
\expandafter\ifx\csname natexlab\endcsname\relax\def\natexlab#1{#1}\fi

\bibitem[{Abu~Sheikha and Inkpen(2011)}]{AbuI2011}
Fadi Abu~Sheikha and Diana Inkpen. 2011.
\newblock \href {https://www.aclweb.org/anthology/W11-2826} {Generation of
  formal and informal sentences}.
\newblock In \emph{Proceedings of the 13th {E}uropean Workshop on Natural
  Language Generation}, pages 187--193, Nancy, France. Association for
  Computational Linguistics.

\bibitem[{Agrawal and Carpuat(2019)}]{AgrawalC2019}
Sweta Agrawal and Marine Carpuat. 2019.
\newblock \href {https://doi.org/10.18653/v1/D19-1166} {Controlling text
  complexity in neural machine translation}.
\newblock In \emph{Proceedings of the 2019 Conference on Empirical Methods in
  Natural Language Processing and the 9th International Joint Conference on
  Natural Language Processing (EMNLP-IJCNLP)}, pages 1549--1564, Hong Kong,
  China. Association for Computational Linguistics.

\bibitem[{Anderson et~al.(2017)Anderson, Fernando, Johnson, and
  Gould}]{AndersonFJG2017}
Peter Anderson, Basura Fernando, Mark Johnson, and Stephen Gould. 2017.
\newblock \href {https://doi.org/10.18653/v1/D17-1098} {Guided open vocabulary
  image captioning with constrained beam search}.
\newblock In \emph{Proceedings of the 2017 Conference on Empirical Methods in
  Natural Language Processing}, pages 936--945, Copenhagen, Denmark.
  Association for Computational Linguistics.

\bibitem[{Arthur et~al.(2016)Arthur, Neubig, and Nakamura}]{Arthur2016}
Philip Arthur, Graham Neubig, and Satoshi Nakamura. 2016.
\newblock \href {https://doi.org/10.18653/v1/D16-1162} {Incorporating discrete
  translation lexicons into neural machine translation}.
\newblock In \emph{Proceedings of the 2016 Conference on Empirical Methods in
  Natural Language Processing}, pages 1557--1567, Austin, Texas. Association
  for Computational Linguistics.

\bibitem[{Bahdanau et~al.(2015)Bahdanau, Cho, and Bengio}]{BahdanauCB15}
Dzmitry Bahdanau, Kyunghyun Cho, and Yoshua Bengio. 2015.
\newblock \href {https://arxiv.org/abs/1409.0473} {Neural machine translation
  by jointly learning to align and translate}.
\newblock In \emph{Proceedings of the 3th International Conference on Learning
  Representations}.

\bibitem[{Bangalore et~al.(2007)Bangalore, Haffner, and
  Kanthak}]{BangaloreHK2007}
Srinivas Bangalore, Patrick Haffner, and Stephan Kanthak. 2007.
\newblock \href {https://www.aclweb.org/anthology/P07-1020} {Statistical
  machine translation through global lexical selection and sentence
  reconstruction}.
\newblock In \emph{Proceedings of the 45th Annual Meeting of the Association of
  Computational Linguistics}, pages 152--159, Prague, Czech Republic.
  Association for Computational Linguistics.

\bibitem[{Barrachina et~al.(2009)Barrachina, Bender, Casacuberta, Civera,
  Cubel, Khadivi, Lagarda, Ney, Tom{\'a}s, Vidal, and Vilar}]{Barrachina2009}
Sergio Barrachina, Oliver Bender, Francisco Casacuberta, Jorge Civera, Elsa
  Cubel, Shahram Khadivi, Antonio Lagarda, Hermann Ney, Jes{\'u}s Tom{\'a}s,
  Enrique Vidal, and Juan-Miguel Vilar. 2009.
\newblock \href {https://doi.org/10.1162/coli.2008.07-055-R2-06-29}
  {Statistical approaches to computer-assisted translation}.
\newblock \emph{Computational Linguistics}, 35(1):3--28.

\bibitem[{Bojar et~al.(2014)Bojar, Buck, Federmann, Haddow, Koehn, Leveling,
  Monz, Pecina, Post, Saint-Amand, Soricut, Specia, and
  Tamchyna}]{BojarWMT2014}
Ond{\v{r}}ej Bojar, Christian Buck, Christian Federmann, Barry Haddow, Philipp
  Koehn, Johannes Leveling, Christof Monz, Pavel Pecina, Matt Post, Herve
  Saint-Amand, Radu Soricut, Lucia Specia, and Ale{\v{s}} Tamchyna. 2014.
\newblock \href {https://doi.org/10.3115/v1/W14-3302} {Findings of the 2014
  workshop on statistical machine translation}.
\newblock In \emph{Proceedings of the Ninth Workshop on Statistical Machine
  Translation}, pages 12--58, Baltimore, Maryland, USA. Association for
  Computational Linguistics.

\bibitem[{Bojar et~al.(2017)Bojar, Chatterjee, Federmann, Graham, Haddow,
  Huang, Huck, Koehn, Liu, Logacheva, Monz, Negri, Post, Rubino, Specia, and
  Turchi}]{Bojar2017WMT}
Ond{\v{r}}ej Bojar, Rajen Chatterjee, Christian Federmann, Yvette Graham, Barry
  Haddow, Shujian Huang, Matthias Huck, Philipp Koehn, Qun Liu, Varvara
  Logacheva, Christof Monz, Matteo Negri, Matt Post, Raphael Rubino, Lucia
  Specia, and Marco Turchi. 2017.
\newblock \href {https://doi.org/10.18653/v1/W17-4717} {Findings of the 2017
  conference on machine translation ({WMT}17)}.
\newblock In \emph{Proceedings of the Second Conference on Machine
  Translation}, pages 169--214, Copenhagen, Denmark. Association for
  Computational Linguistics.

\bibitem[{Bojar et~al.(2016)Bojar, Chatterjee, Federmann, Graham, Haddow, Huck,
  Jimeno~Yepes, Koehn, Logacheva, Monz, Negri, N{\'e}v{\'e}ol, Neves, Popel,
  Post, Rubino, Scarton, Specia, Turchi, Verspoor, and Zampieri}]{BojarWMT2016}
Ond{\v{r}}ej Bojar, Rajen Chatterjee, Christian Federmann, Yvette Graham, Barry
  Haddow, Matthias Huck, Antonio Jimeno~Yepes, Philipp Koehn, Varvara
  Logacheva, Christof Monz, Matteo Negri, Aur{\'e}lie N{\'e}v{\'e}ol, Mariana
  Neves, Martin Popel, Matt Post, Raphael Rubino, Carolina Scarton, Lucia
  Specia, Marco Turchi, Karin Verspoor, and Marcos Zampieri. 2016.
\newblock \href {https://doi.org/10.18653/v1/W16-2301} {Findings of the 2016
  conference on machine translation}.
\newblock In \emph{Proceedings of the First Conference on Machine Translation:
  Volume 2, Shared Task Papers}, pages 131--198, Berlin, Germany. Association
  for Computational Linguistics.

\bibitem[{Brown et~al.(1990)Brown, Cocke, Della~Pietra, Della~Pietra, Jelinek,
  Lafferty, Mercer, and Roossin}]{Brown1990SMT}
Peter~F. Brown, John Cocke, Stephen~A. Della~Pietra, Vincent~J. Della~Pietra,
  Fredrick Jelinek, John~D. Lafferty, Robert~L. Mercer, and Paul~S. Roossin.
  1990.
\newblock \href {https://www.aclweb.org/anthology/J90-2002} {A statistical
  approach to machine translation}.
\newblock \emph{Computational Linguistics}, 16(2):79--85.

\bibitem[{Cheng et~al.(2018)Cheng, Yan, Wagener, and Boots}]{ChengYWB2018}
Ching-An Cheng, Xinyan Yan, Nolan Wagener, and Byron Boots. 2018.
\newblock \href {http://auai.org/uai2018/proceedings/papers/302.pdf} {Fast
  policy learning through imitation and reinforcement}.
\newblock In \emph{Proceedings of the 2018 Conference on Uncertainty in
  Artificial Intelligence (UAI)}, pages 845--855, Monterey, CA, USA.

\bibitem[{Cho et~al.(2014)Cho, van Merri{\"e}nboer, Bahdanau, and
  Bengio}]{Cho2014}
Kyunghyun Cho, Bart van Merri{\"e}nboer, Dzmitry Bahdanau, and Yoshua Bengio.
  2014.
\newblock \href {https://doi.org/10.3115/v1/W14-4012} {On the properties of
  neural machine translation: Encoder{--}decoder approaches}.
\newblock In \emph{Proceedings of {SSST}-8, Eighth Workshop on Syntax,
  Semantics and Structure in Statistical Translation}, pages 103--111, Doha,
  Qatar. Association for Computational Linguistics.

\bibitem[{Chorowski et~al.(2015)Chorowski, Bahdanau, Serdyuk, Cho, and
  Bengio}]{Chorowski2015}
Jan~K Chorowski, Dzmitry Bahdanau, Dmitriy Serdyuk, Kyunghyun Cho, and Yoshua
  Bengio. 2015.
\newblock \href
  {https://papers.nips.cc/paper/2015/file/1068c6e4c8051cfd4e9ea8072e3189e2-Paper.pdf}
  {Attention-based models for speech recognition}.
\newblock In \emph{Advances in neural information processing systems}, pages
  577--585, Montreal, Canada.

\bibitem[{Clark et~al.(2011)Clark, Dyer, Lavie, and Smith}]{Clark2011}
Jonathan~H. Clark, Chris Dyer, Alon Lavie, and Noah~A. Smith. 2011.
\newblock \href {https://www.aclweb.org/anthology/P11-2031} {Better hypothesis
  testing for statistical machine translation: Controlling for optimizer
  instability}.
\newblock In \emph{Proceedings of the 49th Annual Meeting of the Association
  for Computational Linguistics: Human Language Technologies}, pages 176--181,
  Portland, Oregon, USA. Association for Computational Linguistics.

\bibitem[{{Daum\'e III} et~al.(2009){Daum\'e III}, Langford, and
  Marcu}]{DaumeLM2009}
Hal {Daum\'e III}, John Langford, and Daniel Marcu. 2009.
\newblock \href {http://users.umiacs.umd.edu/~hal/docs/daume09searn.pdf}
  {Search-based structured prediction}.
\newblock \emph{Machine learning}, 75(3):297--325.

\bibitem[{Dinu et~al.(2019)Dinu, Mathur, Federico, and
  Al-Onaizan}]{DinuMFA2019}
Georgiana Dinu, Prashant Mathur, Marcello Federico, and Yaser Al-Onaizan. 2019.
\newblock \href {https://doi.org/10.18653/v1/P19-1294} {Training neural machine
  translation to apply terminology constraints}.
\newblock In \emph{Proceedings of the 57th Annual Meeting of the Association
  for Computational Linguistics}, pages 3063--3068, Florence, Italy.
  Association for Computational Linguistics.

\bibitem[{Durrani et~al.(2015)Durrani, Schmid, Fraser, Koehn, and
  Sch{\"u}tze}]{DurraniSFKS2015}
Nadir Durrani, Helmut Schmid, Alexander Fraser, Philipp Koehn, and Hinrich
  Sch{\"u}tze. 2015.
\newblock \href {https://doi.org/10.1162/COLI_a_00218} {The operation sequence
  {M}odel{---}{C}ombining n-gram-based and phrase-based statistical machine
  translation}.
\newblock \emph{Computational Linguistics}, 41(2):157--186.

\bibitem[{Ficler and Goldberg(2017)}]{FiclerG2017}
Jessica Ficler and Yoav Goldberg. 2017.
\newblock \href {https://doi.org/10.18653/v1/W17-4912} {Controlling linguistic
  style aspects in neural language generation}.
\newblock In \emph{Proceedings of the Workshop on Stylistic Variation}, pages
  94--104, Copenhagen, Denmark. Association for Computational Linguistics.

\bibitem[{Foster et~al.(2002)Foster, Langlais, and Lapalme}]{Foster2002}
George Foster, Philippe Langlais, and Guy Lapalme. 2002.
\newblock \href {https://doi.org/10.3115/1118693.1118713} {User-friendly text
  prediction for translators}.
\newblock In \emph{Proceedings of the 2002 Conference on Empirical Methods in
  Natural Language Processing ({EMNLP} 2002)}, pages 148--155. Association for
  Computational Linguistics.

\bibitem[{Ghazvininejad et~al.(2019)Ghazvininejad, Levy, Liu, and
  Zettlemoyer}]{GhazvininejadLLZ2019}
Marjan Ghazvininejad, Omer Levy, Yinhan Liu, and Luke Zettlemoyer. 2019.
\newblock \href {https://doi.org/10.18653/v1/D19-1633} {Mask-predict: Parallel
  decoding of conditional masked language models}.
\newblock In \emph{Proceedings of the 2019 Conference on Empirical Methods in
  Natural Language Processing and the 9th International Joint Conference on
  Natural Language Processing (EMNLP-IJCNLP)}, pages 6112--6121, Hong Kong,
  China. Association for Computational Linguistics.

\bibitem[{Gu et~al.(2018)Gu, Bradbury, Xiong, Li, and Socher}]{GuBXLS2018}
Jiatao Gu, James Bradbury, Caiming Xiong, Victor~OK Li, and Richard Socher.
  2018.
\newblock \href {https://openreview.net/forum?id=B1l8BtlCb} {Non-autoregressive
  neural machine translation}.
\newblock In \emph{International Conference on Learning Representations}.

\bibitem[{Gu et~al.(2019)Gu, Wang, and Zhao}]{GuWZ2019}
Jiatao Gu, Changhan Wang, and Junbo Zhao. 2019.
\newblock \href {http://papers.nips.cc/paper/9297-levenshtein-transformer.pdf}
  {Levenshtein transformer}.
\newblock In \emph{Advances in Neural Information Processing Systems 32}, pages
  11181--11191. Curran Associates, Inc.

\bibitem[{Hieber et~al.(2017)Hieber, Domhan, Denkowski, Vilar, Sokolov,
  Clifton, and Post}]{sockeye2017}
Felix Hieber, Tobias Domhan, Michael Denkowski, David Vilar, Artem Sokolov, Ann
  Clifton, and Matt Post. 2017.
\newblock \href {http://arxiv.org/abs/1712.05690} {Sockeye: {A} toolkit for
  neural machine translation}.
\newblock \emph{CoRR}, abs/1712.05690.

\bibitem[{Hokamp and Liu(2017)}]{HokampL2017}
Chris Hokamp and Qun Liu. 2017.
\newblock \href {https://doi.org/10.18653/v1/P17-1141} {Lexically constrained
  decoding for sequence generation using grid beam search}.
\newblock In \emph{Proceedings of the 55th Annual Meeting of the Association
  for Computational Linguistics (Volume 1: Long Papers)}, pages 1535--1546,
  Vancouver, Canada. Association for Computational Linguistics.

\bibitem[{Isozaki et~al.(2010)Isozaki, Hirao, Duh, Sudoh, and
  Tsukada}]{Isozaki2010RIBES}
Hideki Isozaki, Tsutomu Hirao, Kevin Duh, Katsuhito Sudoh, and Hajime Tsukada.
  2010.
\newblock \href {https://www.aclweb.org/anthology/D10-1092} {Automatic
  evaluation of translation quality for distant language pairs}.
\newblock In \emph{Proceedings of the 2010 Conference on Empirical Methods in
  Natural Language Processing}, pages 944--952, Cambridge, MA. Association for
  Computational Linguistics.

\bibitem[{Kasai et~al.(2020)Kasai, Pappas, Peng, Cross, and
  Smith}]{KasaiPPCS2020}
Jungo Kasai, Nikolaos Pappas, Hao Peng, James Cross, and Noah~A Smith. 2020.
\newblock \href {https://arxiv.org/abs/2006.10369} {Deep encoder, shallow
  decoder: Reevaluating the speed-quality tradeoff in machine translation}.
\newblock \emph{arXiv preprint arXiv:2006.10369}.

\bibitem[{Kingma and Ba(2015)}]{KingmaB15}
Diederik~P. Kingma and Jimmy Ba. 2015.
\newblock \href {https://arxiv.org/pdf/1412.6980.pdf} {Adam: A method for
  stochastic optimization}.
\newblock In \emph{Proceedings of the 3th International Conference on Learning
  Representations}, San Diego, CA, USA.

\bibitem[{Koehn et~al.(2007)Koehn, Hoang, Birch, Callison-Burch, Federico,
  Bertoldi, Cowan, Shen, Moran, Zens, Dyer, Bojar, Constantin, and
  Herbst}]{Koehn2007Moses}
Philipp Koehn, Hieu Hoang, Alexandra Birch, Chris Callison-Burch, Marcello
  Federico, Nicola Bertoldi, Brooke Cowan, Wade Shen, Christine Moran, Richard
  Zens, Chris Dyer, Ond{\v{r}}ej Bojar, Alexandra Constantin, and Evan Herbst.
  2007.
\newblock \href {https://www.aclweb.org/anthology/P07-2045} {{M}oses: Open
  source toolkit for statistical machine translation}.
\newblock In \emph{Proceedings of the 45th Annual Meeting of the Association
  for Computational Linguistics Companion Volume Proceedings of the Demo and
  Poster Sessions}, pages 177--180, Prague, Czech Republic. Association for
  Computational Linguistics.

\bibitem[{Kudo and Richardson(2018)}]{KudoR2018}
Taku Kudo and John Richardson. 2018.
\newblock \href {https://doi.org/10.18653/v1/D18-2012} {{S}entence{P}iece: A
  simple and language independent subword tokenizer and detokenizer for neural
  text processing}.
\newblock In \emph{Proceedings of the 2018 Conference on Empirical Methods in
  Natural Language Processing: System Demonstrations}, pages 66--71, Brussels,
  Belgium. Association for Computational Linguistics.

\bibitem[{Lample et~al.(2018)Lample, Conneau, Denoyer, and
  Ranzato}]{LampleCDR18}
Guillaume Lample, Alexis Conneau, Ludovic Denoyer, and Marc{'}Aurelio Ranzato.
  2018.
\newblock \href {https://openreview.net/forum?id=rkYTTf-AZ} {Unsupervised
  machine translation using monolingual corpora only}.
\newblock In \emph{Proceedings of the 6th International Conference on Learning
  Representations}.

\bibitem[{Leblond et~al.(2018)Leblond, Alayrac, Osokin, and
  Lacoste-Julien}]{LeblondAOL18}
R{\'{e}}mi Leblond, Jean-Baptiste Alayrac, Anton Osokin, and Simon
  Lacoste-Julien. 2018.
\newblock \href {https://openreview.net/forum?id=HkUR_y-RZ} {{SEARNN}: Training
  {RNN}s with global-local losses}.
\newblock In \emph{International Conference on Learning Representations}.

\bibitem[{Lee et~al.(2018)Lee, Mansimov, and Cho}]{LeeMC2018}
Jason Lee, Elman Mansimov, and Kyunghyun Cho. 2018.
\newblock \href {https://doi.org/10.18653/v1/D18-1149} {Deterministic
  non-autoregressive neural sequence modeling by iterative refinement}.
\newblock In \emph{Proceedings of the 2018 Conference on Empirical Methods in
  Natural Language Processing}, pages 1173--1182, Brussels, Belgium.
  Association for Computational Linguistics.

\bibitem[{Levenshtein(1966)}]{Levenshtein1966}
Vladimir~I Levenshtein. 1966.
\newblock \href {https://ui.adsabs.harvard.edu/abs/1966SPhD...10..707L} {Binary
  codes capable of correcting deletions, insertions and reversals}.
\newblock \emph{Soviet Physics Doklady}, 10:707.

\bibitem[{Ma et~al.(2019)Ma, Zhou, Li, Neubig, and Hovy}]{MaZLNH2019}
Xuezhe Ma, Chunting Zhou, Xian Li, Graham Neubig, and Eduard Hovy. 2019.
\newblock \href {https://doi.org/10.18653/v1/D19-1437} {{F}low{S}eq:
  Non-autoregressive conditional sequence generation with generative flow}.
\newblock In \emph{Proceedings of the 2019 Conference on Empirical Methods in
  Natural Language Processing and the 9th International Joint Conference on
  Natural Language Processing (EMNLP-IJCNLP)}, pages 4282--4292, Hong Kong,
  China. Association for Computational Linguistics.

\bibitem[{Mei et~al.(2016)Mei, Bansal, and Walter}]{MeiBW2016}
Hongyuan Mei, Mohit Bansal, and Matthew~R. Walter. 2016.
\newblock \href {https://doi.org/10.18653/v1/N16-1086} {What to talk about and
  how? selective generation using {LSTM}s with coarse-to-fine alignment}.
\newblock In \emph{Proceedings of the 2016 Conference of the North {A}merican
  Chapter of the Association for Computational Linguistics: Human Language
  Technologies}, pages 720--730, San Diego, California. Association for
  Computational Linguistics.

\bibitem[{Nakazawa et~al.(2017)Nakazawa, Higashiyama, Ding, Mino, Goto, Kazawa,
  Oda, Neubig, and Kurohashi}]{NakazawaWAT2017}
Toshiaki Nakazawa, Shohei Higashiyama, Chenchen Ding, Hideya Mino, Isao Goto,
  Hideto Kazawa, Yusuke Oda, Graham Neubig, and Sadao Kurohashi. 2017.
\newblock \href {https://www.aclweb.org/anthology/W17-5701} {Overview of the
  4th workshop on {A}sian translation}.
\newblock In \emph{Proceedings of the 4th Workshop on {A}sian Translation
  ({WAT}2017)}, pages 1--54, Taipei, Taiwan. Asian Federation of Natural
  Language Processing.

\bibitem[{Neubig et~al.(2019)Neubig, Dou, Hu, Michel, Pruthi, and
  Wang}]{Neubig2019compare}
Graham Neubig, Zi-Yi Dou, Junjie Hu, Paul Michel, Danish Pruthi, and Xinyi
  Wang. 2019.
\newblock \href {https://doi.org/10.18653/v1/N19-4007} {compare-mt: A tool for
  holistic comparison of language generation systems}.
\newblock In \emph{Proceedings of the 2019 Conference of the North {A}merican
  Chapter of the Association for Computational Linguistics (Demonstrations)},
  pages 35--41, Minneapolis, Minnesota. Association for Computational
  Linguistics.

\bibitem[{Nguyen and Chiang(2018)}]{NguyenC18}
Toan~Q. Nguyen and David Chiang. 2018.
\newblock \href {http://aclweb.org/anthology/N18-1031} {Improving lexical
  choice in neural machine translation}.
\newblock In \emph{Proceedings of the 2018 Conference of the North American
  Chapter of the Association for Computational Linguistics: Human Language
  Technologies}, pages 334--343. Association for Computational Linguistics.

\bibitem[{van~den Oord et~al.(2018)van~den Oord, Li, Babuschkin, Simonyan,
  Vinyals, Kavukcuoglu, van~den Driessche, Lockhart, Cobo, Stimberg,
  Casagrande, Grewe, Noury, Dieleman, Elsen, Kalchbrenner, Zen, Graves, King,
  Walters, Belov, and Hassabis}]{Oord2018}
Aaron van~den Oord, Yazhe Li, Igor Babuschkin, Karen Simonyan, Oriol Vinyals,
  Koray Kavukcuoglu, George van~den Driessche, Edward Lockhart, Luis Cobo,
  Florian Stimberg, Norman Casagrande, Dominik Grewe, Seb Noury, Sander
  Dieleman, Erich Elsen, Nal Kalchbrenner, Heiga Zen, Alex Graves, Helen King,
  Tom Walters, Dan Belov, and Demis Hassabis. 2018.
\newblock \href {http://proceedings.mlr.press/v80/oord18a.html} {Parallel
  {W}ave{N}et: Fast high-fidelity speech synthesis}.
\newblock In \emph{Proceedings of the 35th International Conference on Machine
  Learning}, volume~80 of \emph{Proceedings of Machine Learning Research},
  pages 3918--3926, Stockholmsmässan, Stockholm Sweden. PMLR.

\bibitem[{Ott et~al.(2019)Ott, Edunov, Baevski, Fan, Gross, Ng, Grangier, and
  Auli}]{Ott2019fairseq}
Myle Ott, Sergey Edunov, Alexei Baevski, Angela Fan, Sam Gross, Nathan Ng,
  David Grangier, and Michael Auli. 2019.
\newblock \href {https://doi.org/10.18653/v1/N19-4009} {Fairseq: A fast,
  extensible toolkit for sequence modeling}.
\newblock In \emph{Proceedings of the 2019 Conference of the North {A}merican
  Chapter of the Association for Computational Linguistics (Demonstrations)},
  pages 48--53, Minneapolis, Minnesota. Association for Computational
  Linguistics.

\bibitem[{Papineni et~al.(2002)Papineni, Roukos, Ward, and
  Zhu}]{Papineni2002BLEU}
Kishore Papineni, Salim Roukos, Todd Ward, and Wei-Jing Zhu. 2002.
\newblock \href {https://doi.org/10.3115/1073083.1073135} {{BLEU}: a method for
  automatic evaluation of machine translation}.
\newblock In \emph{Proceedings of the 40th Annual Meeting of the Association
  for Computational Linguistics}, pages 311--318, Philadelphia, Pennsylvania,
  USA. Association for Computational Linguistics.

\bibitem[{Post and Vilar(2018)}]{PostV2018}
Matt Post and David Vilar. 2018.
\newblock \href {https://doi.org/10.18653/v1/N18-1119} {Fast lexically
  constrained decoding with dynamic beam allocation for neural machine
  translation}.
\newblock In \emph{Proceedings of the 2018 Conference of the North {A}merican
  Chapter of the Association for Computational Linguistics: Human Language
  Technologies, Volume 1 (Long Papers)}, pages 1314--1324, New Orleans,
  Louisiana. Association for Computational Linguistics.

\bibitem[{Press and Wolf(2017)}]{PressW17}
Ofir Press and Lior Wolf. 2017.
\newblock \href {http://aclweb.org/anthology/E17-2025} {Using the output
  embedding to improve language models}.
\newblock In \emph{Proceedings of the 15th Conference of the European Chapter
  of the Association for Computational Computational}, pages 157--163.
  Association for Computational Linguistics.

\bibitem[{Ross and Bagnell(2014)}]{RossB2014}
St{\'{e}}phane Ross and J.~Andrew Bagnell. 2014.
\newblock \href {http://arxiv.org/abs/1406.5979} {Reinforcement and imitation
  learning via interactive no-regret learning}.
\newblock \emph{CoRR}, abs/1406.5979.

\bibitem[{Ross et~al.(2011)Ross, Gordon, and Bagnell}]{RossGB11}
Stephane Ross, Geoffrey Gordon, and Drew Bagnell. 2011.
\newblock \href {http://proceedings.mlr.press/v15/ross11a.html} {A reduction of
  imitation learning and structured prediction to no-regret online learning}.
\newblock In \emph{Proceedings of the Fourteenth International Conference on
  Artificial Intelligence and Statistics}, volume~15 of \emph{Proceedings of
  Machine Learning Research}, pages 627--635. PMLR.

\bibitem[{Sennrich et~al.(2016{\natexlab{a}})Sennrich, Haddow, and
  Birch}]{SennrichHB2016politedness}
Rico Sennrich, Barry Haddow, and Alexandra Birch. 2016{\natexlab{a}}.
\newblock \href {https://doi.org/10.18653/v1/N16-1005} {Controlling politeness
  in neural machine translation via side constraints}.
\newblock In \emph{Proceedings of the 2016 Conference of the North {A}merican
  Chapter of the Association for Computational Linguistics: Human Language
  Technologies}, pages 35--40, San Diego, California. Association for
  Computational Linguistics.

\bibitem[{Sennrich et~al.(2016{\natexlab{b}})Sennrich, Haddow, and
  Birch}]{SennrichHB16bpe}
Rico Sennrich, Barry Haddow, and Alexandra Birch. 2016{\natexlab{b}}.
\newblock \href {http://www.aclweb.org/anthology/P16-1162} {Neural machine
  translation of rare words with subword units}.
\newblock In \emph{Proceedings of the 54th Annual Meeting of the Association
  for Computational Linguistics}, pages 1715--1725. Association for
  Computational Linguistics.

\bibitem[{Song et~al.(2019)Song, Zhang, Yu, Luo, Wang, and
  Zhang}]{SongZYLWZ2019}
Kai Song, Yue Zhang, Heng Yu, Weihua Luo, Kun Wang, and Min Zhang. 2019.
\newblock \href {https://doi.org/10.18653/v1/N19-1044} {Code-switching for
  enhancing {NMT} with pre-specified translation}.
\newblock In \emph{Proceedings of the 2019 Conference of the North {A}merican
  Chapter of the Association for Computational Linguistics: Human Language
  Technologies, Volume 1 (Long and Short Papers)}, pages 449--459, Minneapolis,
  Minnesota. Association for Computational Linguistics.

\bibitem[{Stahlberg et~al.(2018)Stahlberg, Saunders, and
  Byrne}]{StahlbergSB2018}
Felix Stahlberg, Danielle Saunders, and Bill Byrne. 2018.
\newblock \href {https://doi.org/10.18653/v1/W18-5420} {An operation sequence
  model for explainable neural machine translation}.
\newblock In \emph{Proceedings of the 2018 {EMNLP} Workshop {B}lackbox{NLP}:
  Analyzing and Interpreting Neural Networks for {NLP}}, pages 175--186,
  Brussels, Belgium. Association for Computational Linguistics.

\bibitem[{Stern et~al.(2019)Stern, Chan, Kiros, and Uszkoreit}]{SternCKU2019}
Mitchell Stern, William Chan, Jamie Kiros, and Jakob Uszkoreit. 2019.
\newblock \href {http://proceedings.mlr.press/v97/stern19a.html} {Insertion
  transformer: Flexible sequence generation via insertion operations}.
\newblock In \emph{Proceedings of the 36th International Conference on Machine
  Learning}, volume~97 of \emph{Proceedings of Machine Learning Research},
  pages 5976--5985, Long Beach, California, USA. PMLR.

\bibitem[{Stern et~al.(2018)Stern, Shazeer, and Uszkoreit}]{SternSU2018}
Mitchell Stern, Noam Shazeer, and Jakob Uszkoreit. 2018.
\newblock \href
  {https://proceedings.neurips.cc/paper/2018/file/c4127b9194fe8562c64dc0f5bf2c93bc-Paper.pdf}
  {Blockwise parallel decoding for deep autoregressive models}.
\newblock In \emph{Advances in Neural Information Processing Systems},
  volume~31, pages 10086--10095, Montreal, Canada. Curran Associates, Inc.

\bibitem[{Susanto et~al.(2020)Susanto, Chollampatt, and Tan}]{SusantoCT2020}
Raymond~Hendy Susanto, Shamil Chollampatt, and Liling Tan. 2020.
\newblock \href {https://www.aclweb.org/anthology/2020.acl-main.325} {Lexically
  constrained neural machine translation with {L}evenshtein transformer}.
\newblock In \emph{Proceedings of the 58th Annual Meeting of the Association
  for Computational Linguistics}, pages 3536--3543, Online. Association for
  Computational Linguistics.

\bibitem[{Tang et~al.(2016)Tang, Meng, Lu, Li, and Yu}]{Tang2016}
Yaohua Tang, Fandong Meng, Zhengdong Lu, Hang Li, and Philip L.~H. Yu. 2016.
\newblock \href {http://arxiv.org/abs/1606.01792} {Neural machine translation
  with external phrase memory}.
\newblock \emph{CoRR}, abs/1606.01792.

\bibitem[{Vaswani et~al.(2017)Vaswani, Shazeer, Parmar, Uszkoreit, Jones,
  Gomez, Kaiser, and Polosukhin}]{Vaswani2017}
Ashish Vaswani, Noam Shazeer, Niki Parmar, Jakob Uszkoreit, Llion Jones,
  Aidan~N Gomez, \L~ukasz Kaiser, and Illia Polosukhin. 2017.
\newblock \href
  {https://proceedings.neurips.cc/paper/2017/file/3f5ee243547dee91fbd053c1c4a845aa-Paper.pdf}
  {Attention is all you need}.
\newblock In \emph{Advances in Neural Information Processing Systems},
  volume~30, pages 5998--6008, Long Beach, CA, USA. Curran Associates, Inc.

\bibitem[{Vinyals and Le(2015)}]{Vinyals2015}
Oriol Vinyals and Quoc Le. 2015.
\newblock \href {https://arxiv.org/abs/1506.05869} {A neural conversational
  model}.
\newblock In \emph{ICML Deep Learning Workshop}, Lille, France.

\bibitem[{Wang et~al.(2018)Wang, Zhang, and Chen}]{WangZC2018}
Chunqi Wang, Ji~Zhang, and Haiqing Chen. 2018.
\newblock \href {https://doi.org/10.18653/v1/D18-1044} {Semi-autoregressive
  neural machine translation}.
\newblock In \emph{Proceedings of the 2018 Conference on Empirical Methods in
  Natural Language Processing}, pages 479--488, Brussels, Belgium. Association
  for Computational Linguistics.

\bibitem[{Wang et~al.(2019)Wang, Tian, He, Qin, Zhai, and Liu}]{WangTHQZL2019}
Yiren Wang, Fei Tian, Di~He, Tao Qin, ChengXiang Zhai, and Tie-Yan Liu. 2019.
\newblock \href {https://doi.org/10.1609/aaai.v33i01.33015377}
  {Non-autoregressive machine translation with auxiliary regularization}.
\newblock \emph{Proceedings of the AAAI Conference on Artificial Intelligence},
  33(01):5377--5384.

\bibitem[{Welleck et~al.(2019)Welleck, Brantley, {Daum\'e III}, and
  Cho}]{WelleckBDC2019}
Sean Welleck, Kiant{\'e} Brantley, Hal {Daum\'e III}, and Kyunghyun Cho. 2019.
\newblock \href {http://proceedings.mlr.press/v97/welleck19a.html}
  {Non-monotonic sequential text generation}.
\newblock In \emph{Proceedings of the 36th International Conference on Machine
  Learning}, volume~97 of \emph{Proceedings of Machine Learning Research},
  pages 6716--6726. PMLR.

\bibitem[{Yvon and Abdul~Rauf(2020)}]{YvonS2020}
Fran{\c c}ois Yvon and Sadaf Abdul~Rauf. 2020.
\newblock \href {https://hal.archives-ouvertes.fr/hal-02895535} {{Utilisation
  de ressources lexicales et terminologiques en traduction neuronale}}.
\newblock Research Report 2020-001, {LIMSI-CNRS}.

\end{thebibliography}
\bibliographystyle{acl_natbib}

\end{document}